%% file: main.tex
\documentclass[sigconf, nonacm]{acmart}
\setkeys{acmart.cls}{balance=false}

\usepackage{graphicx}
\usepackage{booktabs}
\usepackage{subcaption}
\usepackage{multirow}
\usepackage{array}
\usepackage{makecell}
\usepackage{amsmath}
\usepackage{algorithm}
\usepackage{algpseudocode}
\usepackage{enumitem}
\graphicspath{{figures/}}
\title{Co-Evolving Behavioral Agents and Detectors for Adaptive Information Operations}

\author{Lin Tian and Marian-Andrei Rizoiu}
\affiliation{%
  \institution{Behavioral Data Science, University of Technology Sydney}
  \city{Sydney}
  \country{Australia}}
\email{\{Lin.Tian-3, Marian-Andrei.Rizoiu\}@uts.edu.au}
\renewcommand{\shortauthors}{Tian and Rizoiu}

\begin{document}

\input{sections/abstract}

\maketitle

\input{sections/intro_v2}

\input{sections/related_work_v2}

\input{sections/method_v5}

\input{sections/exp_v5}

\input{sections/case_study_v8}

\input{sections/conclusion_v3}

\bibliographystyle{ACM-Reference-Format}
\bibliography{ref}

\input{sections/appendix}

\end{document}

%% file: sections/abstract.tex
\begin{abstract}
Detectors of state-sponsored information operations (IOs) are trained on campaigns that platforms have already taken down.
They lose much of their accuracy on an actor they have not seen, even though no operator in the archived data is trying to avoid them.
A real operator wants two things, to reach an audience and to avoid being caught.
We present IO-AI, a framework that gives simulated agents these two capabilities, engagement and evasion, and tests what each one does to detection.
The agents are built from the posting behavior of real campaigns and choose the emotion, type, and timing of each post.
An agent and a detector then train against each other in turn.
We rank the trained agents by how hard they are to detect and train one detector on them from easiest to hardest.
On 20 Twitter IO campaigns, this detector reduces missed detections of adaptive agents by $65\%$ relative to a detector trained only on the archived campaigns, at a similar false-alarm rate.
On actors it has never seen it reaches an $F_1$ of $0.81$, against $0.62$.
We find that an agent trained to seek engagement becomes easier to catch, because it learns the fast, emotional retweeting that archived campaigns were caught producing.
An agent that was trained for engagement before the game keeps its engagement lead after it learns to evade.
We also introduce a behavioral distance metric that compares two groups of accounts using only their unlabeled posting histories.
The larger the distance between two groups, the worse a detector trained on one does on the other (Spearman $\rho_s=-0.58$), so the metric can be used to choose which new groups to test first.
\end{abstract}

%% file: sections/intro_v2.tex
\section{Introduction}
\label{sec:intro}
\input{figs/fig_cross_actor}
Following the 2016 U.S. election, Twitter (now X) identified 3,841 accounts affiliated with Russia's Internet Research Agency (IRA) in a retrospective archive spanning multiple years~\citep{gadde2018enabling}\footnote{\url{https://blog.x.com/en_us/topics/company/2018/enabling-further-research-of-information-operations-on-twitter}}. 
These accounts adopted different roles, including political commentators and local news personas~\citep{howard2018ira}.\footnote{\url{https://www.intelligence.senate.gov/wp-content/uploads/2024/08/sites-default-files-documents-the-ira-social-media-and-political-polarization.pdf}}. 
In December 2021, Twitter disclosed a separate 16-account network linked to the IRA that attempted to introduce pro-Russia viewpoints into political discussion in the Central African Republic~\citep{twitter2021car}\footnote{\url{https://blog.x.com/en_us/topics/company/2021/disclosing-state-linked-information-operations-we-ve-removed}}. 
Such archives make systematic analysis possible, but each records behavior from a particular operation and period.
How can we train detectors to recognize new information operations when their operators adapt to evade detection?

Information operation (IO) detectors are built and evaluated on archived campaigns, using the text of the posts, the way the posts spread, the behavior of the accounts behind them, or some combination of these~\citep{alizadeh2020content,liu2018early,varol2017online,luceri2024unmasking,ezzeddine2023exposing,schneider2026beyond,minici2025iohunter}.
Figure~\ref{fig:cross_actor} shows how much of what they learn carries from one state actor to another.
Trained and tested on the same actor, detectors built on text, propagation, or their fusion reach a mean $F_1$ of $0.84$ to $0.92$; trained on one actor and tested on another, the same detectors fall to $0.52$ to $0.55$.
The drop arises on archived data, with no operator responding to the detector, so most of what each detector learned was specific to the actor it was trained on.
Few-shot transfer adapts a troll detector to a new campaign from a handful of labels~\citep{tian2023metatroll}, early detectors decide from a partial history~\citep{ma2016detecting}, and adversarial studies train bots to evade a fixed detector~\citep{le2022socialbots} or manipulate the engagement a post receives to fool one~\citep{zhang2023attacking}.
In each of these settings one side stays fixed, whether the detector is updated once a campaign is exposed or the operator learns against a detector that never retrains.
What pursuing engagement, evasion, or both does to an operator when the detector also learns, and what a detector gains from training on the result, remain open.
Operators also want reach, which comes from the people who share a post~\citep{vosoughi2018spread} and from the unwitting participants a campaign recruits to amplify it~\citep{starbird2019disinformation}.
Which strategies survive when operators pursue engagement as well as evasion and the detector learns in turn, and what must a detector train against to counter them?

We introduce IO-AI, a framework for training sequential IO detectors against a population of adaptive behavioral agents (Figure~\ref{fig:framework}).
Agents choose post emotion, posting type, and timing; a frozen text model renders each action into a post.
This compact action space makes policies interpretable and supports cloning from the takedown archives.
We cross three starting policies (cloned, engagement-pretrained, and static-detector-pretrained) with three objectives (evasion, predicted engagement, and both), alongside a fixed cloned control.
In the adaptive games, agents and detectors train in turn, while the detector decides whether to wait, flag, or pass.
Cross-play ranks agents across the learned detector population, giving an easy-to-hard curriculum for a single detector.

Our contributions connect behavioral analysis to detector training. 
First, the game family exposes how initialization and payoff shape adaptation, including the difference between engagement pretraining and engagement optimization during play. 
Second, cross-play supplies an empirical difficulty ordering whose contribution we assess against static training, single-game training, and alternative population schedules. 
Third, policies cloned from cohort histories provide a behavioral-distance diagnostic that we relate to detector transfer. 
Experiments using Twitter campaign data, simulated adversaries, and re-attacks evaluate the agents behavioral coverage.

%% file: figs/fig_cross_actor.tex
\begin{figure*}[t]
\centering
\includegraphics[width=\textwidth]{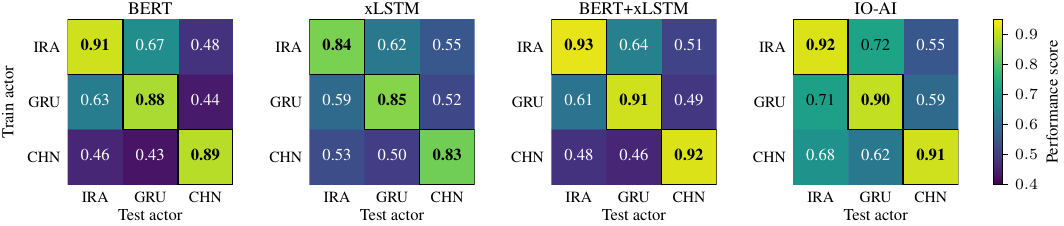}
\caption{
Cross-actor detection performance ($F_1$) of BERT, xLSTM, their fusion (BERT+xLSTM), and IO-AI. Rows indicate training actors, columns indicate test actors, and diagonal cells show in-domain performance.}
\label{fig:cross_actor}
\end{figure*}

%% file: sections/related_work_v2.tex
\section{Related Work}
\label{sec:related}

\paragraph{Information operation and bot detection.}
Prior work on social bot and information operation detection has built strong supervised models from text, account metadata, network structure, and temporal behavior~\citep{ferrara2016rise,varol2017online}.
These systems are useful when future campaigns look like past campaigns, but they are trained against fixed data distributions. 
We extend behavior-based IO detection to adaptive opponents by training sequential detectors against learned adversaries and evaluating the transfer across actors and eras.

\paragraph{Adversarial detection and games.}
Adversarial machine learning studies how classifiers fail when an attacker chooses the inputs~\citep{biggio2013evasion,goodfellow2015explaining}; GANs~\citep{goodfellow2014gan} and PGD-robust training~\citep{madry2018towards} are the standard two-player learning formulations. 
Security games make the interaction explicit by casting attacker and defender as players in a Markov or Stackelberg game~\citep{littman1994markov,sinha2018stackelberg}. 
We adapt this game-theoretic perspective to the sequential behavior of IO agents: the agent controls post emotion, posting type, and timing, while the detector decides whether to wait, flag, or pass.

\paragraph{Empirical game theory and game-space structure.}
Empirical game-theoretic analysis (EGTA) studies strategic interactions using payoffs estimated from simulated policy matchups~\citep{wellman2006methods,tuyls2018generalised,wellman2024egta}. 
Related population-based approaches include policy-space response oracles (PSRO) and league-based self-play~\citep{lanctot2017unified,vinyals2019alphastar}. 
Open-ended-learning work has analyzed the strategic geometry of single games~\citep{balduzzi2019open,czarnecki2020spinning}, while \citet{sokota2023learning} learn payoff models across parameterized game families. 
We take the game theory to IO detection. 
Instead of one game, we solve ten partially observable games between an IO agent and a sequential detector, one for each choice of starting policy and objective, and then play the solved agents against every solved detector. 
The resulting cross-play ordering is what our detector curriculum trains over.

\paragraph{Curriculum learning.}
Curriculum learning orders training from easy cases to hard ones~\citep{bengio2009curriculum}, and league-based self-play produces diverse adaptive opponents~\citep{vinyals2019alphastar}. 
We build on these ideas by using cross-play to measure the difficulty of co-evolved IO strategies and construct a curriculum for training a single sequential detector.

%% file: sections/method_v5.tex
\section{Methodology}
\label{sec:method}
We train detectors to recognize IO accounts as their behavior adapts to detection. 
IO-AI first co-evolves agents and detectors across ten games, then uses the resulting agents to construct a detector curriculum ordered by cross-play miss rate (Figure~\ref{fig:framework}).
\input{figs/fig_framework}

\subsection{Problem Formulation}
\label{sec:formulation}
Let $x_{1:t} = (x_1, \ldots, x_t)$ denote an account's observed history, where each $x_i$ records text, action type, timing, and received engagement.
Each episode has a label $y\in\{0,1\}$, where $y=1$ denotes IO and $y=0$ a matched non-IO control.
Our goal is to identify IO accounts from partial histories while limiting false alarms and detection delay.
We model the IO agent $g$ and detector $d$ using a partially observable Markov game~\citep{littman1994markov,hansen2004dynamic,lanctot2017unified}, $\mathcal{G} = (\mathcal{S}, \mathcal{A}_g, \mathcal{A}_d, P, \Omega, r_g, r_d, \gamma)$.
Here $\mathcal{S}$ is the state space, $\mathcal{A}_g$ and $\mathcal{A}_d$ are the player action spaces, $P$ is the transition kernel, and $\Omega$ is the observation map.
The agent reward $r_g$ is set per game by its objective, the detector reward $r_d$ is the same to all games, and $\gamma \in (0,1]$ is the discount.
The state $s_t \in \mathcal{S}$ contains $y$ and $x_{1:t}$.
Each episode begins from a fixed seed prefix of $t_0$ recorded posts, so the detector's first decision is at $t=t_0$ and all step counts are measured in decision steps.
The detector observes only $o_t = \Omega(s_t) = x_{1:t}$, with $y$ hidden, and chooses $a^d_t \in \mathcal{A}_d = \{\textsc{wait}, \textsc{flag}, \textsc{pass}\}$; throughout, $a_t \in \mathcal{A}_g$ denotes the agent's action and $a^d_t$ the detector's.
The agent observes $s_t$ and whether the detector continues, but not the detector's hidden state, action probabilities, or decision boundary.
The transition $P(s_{t+1}\mid s_t,a_t,a^d_t)$ appends $x_{t+1}$ only after \textsc{wait}.
For $y=1$, $x_{t+1}$ is rendered from $a_t$; for $y=0$, it is replayed from the matched control history.
Both \textsc{flag} and \textsc{pass} terminate the episode at step $T$, which is bounded by the horizon $T_{\max}$, the largest number of decision steps an episode may run.
An episode reaching $T_{\max}$ without a flag defaults to \textsc{pass}. $a^d_T$ denotes the effective terminal decision after this horizon rule.
We summarize terminal outcomes by
\begin{equation}
\begin{aligned}
\mathrm{MissRate} &= \Pr(a^d_T=\textsc{pass}\mid y=1),\\
\mathrm{FalseAlarm} &= \Pr(a^d_T=\textsc{flag}\mid y=0).
\end{aligned}
\label{eq:terminal_error_rates}
\end{equation}

\subsection{Agent Initializations}
\label{sec:agent}
In an IO episode, the agent samples an action $a_t \sim \pi_g(\cdot \mid s_t)$; in every episode, the detector reads the history step by step and decides whether to keep watching.

\textbf{Action space.}
Each action $a_t = (e_t, c_t, \tau_t)$ has three components, read off the raw posts, timestamps, and per-user timelines that define the state: 
the \emph{emotion} $e_t$ of the next post, one of the GoEmotions labels~\citep{demszky2020goemotions}\footnote{We label each observed post with the \texttt{roberta-base-go\_emotions} classifier (\url{https://huggingface.co/SamLowe/roberta-base-go_emotions}), a RoBERTa-base model fine-tuned on the 28 GoEmotions labels, which supplies the cloning target for this head.}; 
the \emph{action type} $c_t \in \{\text{post}, \text{retweet}\}$, 
and the \emph{inter-action interval} $\tau_t$, how long the agent waits after $x_t$ before its next post. The timing head $\pi_\tau$ is categorical over bins of $\tau_t$ that are evenly spaced in $\log_{10}$ minutes and span ten seconds to one day.
The policy is autoregressive over these components:
\begin{equation}
\pi_g(a_t \mid s_t) = \pi_e(e_t \mid s_t)\, \pi_c(c_t \mid s_t, e_t)\, \pi_\tau(\tau_t \mid s_t, e_t, c_t).
\label{eq:autoregressive}
\end{equation}

\textbf{Agent base model.}
The policy $\pi_g$ is a six-block Mamba-2 sequence model~\citep{dao2024transformers} over the history $x_{1:t}$, with one token per post built from the post's emotion label, posting type, timing, received engagement and a frozen text embedding, and three heads that implement Equation~\eqref{eq:autoregressive} (Appendix~\ref{sec:agent_arch_app}). 
A frozen renderer $q_{\text{txt}}(\cdot \mid a_t, \kappa)$, conditioned on a campaign/topic context $\kappa$ fixed for the episode, maps each action to post text, and a deterministic updater appends the matching timing and propagation features to the public history. 

\textbf{Detector architecture.}
\label{sec:detector}
The detector reads a generated IO or replayed control history step by step. An encoder $\phi$ maps the observed prefix to $h_t = \phi(x_{\le t})$ by fusing a text encoder, which follows IC-Mamba~\cite{tian2025icmamba} for the interval-censored input history, with a Mamba-2 sequence model~\cite{dao2024transformers} over action type, timing, and propagation features. 
A recurrent decision module keeps a summary $z_t = f(z_{t-1}, h_t)$, with initial state $z_0$, and the detector samples
$
a^d_t \sim \pi_d(\cdot \mid z_t), a^d_t \in \{\textsc{wait}, \textsc{flag}, \textsc{pass}\}.
$

\textbf{Behavioral cloning (BC).} 
We fit $\pi_g^{\text{BC}}$ to real IO data $\mathcal{D}_{\text{IO}}$ by behavioral cloning~\citep{pomerleau1989alvinn}, minimizing
\begin{equation*}
\mathcal{L}_{\text{BC}} = -\mathbb{E}_{(s,a)\sim\mathcal{D}_{\text{IO}}}\big[\log \pi_g(a \mid s)\big]
\end{equation*}
under the factorization of Equation~\eqref{eq:autoregressive}. This anchors the initial policy to the observed behavior of IO accounts.

\textbf{Engagement Max (EM).}
EM starts from $\pi_g^{\text{BC}}$ and is tuned to prefer actions that a frozen engagement model, $R_{\text{eng}}$, predicts will draw more reactions. 
$R_{\text{eng}}$ is the DREAMS predictor~\citep{tian2026dreams}, which scores a post from its text and propagation signals.\footnote{We fit $R_{\text{eng}}$ on the training split before any adversarial optimization; held-out campaigns and periods never enter its training data.} 
We use direct preference optimization (DPO)~\citep{rafailov2023direct} rather than policy gradients on $R_{\text{eng}}$: it trains offline on a fixed preference set, its implicit KL penalty to the reference policy (scaled by $\beta$) keeps EM near observed IO behavior without a separate regularizer, and it uses $R_{\text{eng}}$ only to order candidate actions, so the score's unit enters only through the margin $\delta$. 
Preference pairs come from BC rollouts on training-split IO prefixes: at each visited state we draw $L$ candidate actions from $\pi_g^{\text{BC}}$, render each with $q_{\text{txt}}$, and score the resulting one-step history with $R_{\text{eng}}$, averaging over $n_{\text{txt}}$ renderer samples. 
We keep every pair $(a^+,a^-)$ whose score gap is at least $\delta$; smaller gaps lie within renderer noise. 
The set $\mathcal{P}_{\text{eng}}$ holds hard-margin preferences rather than Bradley--Terry probabilities~\citep{bradley1952rank}. 
We minimize the DPO objective (Appendix Equation~\eqref{eq:dpo_app}) with $\pi_g^{\text{BC}}$ as reference.

\textbf{Static-detector pretrained (SD).}
SD starts from $\pi_g^{\text{BC}}$ and learns to evade a frozen reference detector, $\pi_d^{\text{ref}}$, before co-evolution begins. 
The reference detector uses Mamba-2~\citep{dao2024transformers} for both its text encoder and its sequence model over posting type, timing and propagation features, followed by the recurrent decision module above; its text encoder therefore differs from the IC-Mamba encoder of the detectors we evaluate. 
We train it by supervised classification on $\mathcal{D}_{\text{IO}}$ and $\mathcal{D}_{\text{ctrl}}$ and freeze its parameters and decision thresholds before SD pretraining begins. 
SD then trains for $N_{\text{SD}}$ iterations under the learning rule, with the Evade reward (Equation~\ref{eq:evade}) and no KL penalty. 
We hold $\pi_d^{\text{ref}}$ out of every detector we evaluate.

\textbf{Control trajectory pool.}
The control pool $\mathcal{D}_{\text{ctrl}}$ contains real non-IO Twitter histories matched to IO campaigns by topic and activity volume. Control trajectories follow the same campaign and time splits as IO trajectories.

\subsection{Game Setup}
\label{sec:games}
Each game pairs an initialization from $\{\text{BC}, \text{EM}, \text{SD}\}$ with an objective $o \in \{\text{Evade}, \text{Engage}, \text{E+E}, \text{Real}\}$, where E+E abbreviates Evade+Engage and Real is paired with BC only. 
We denote the resulting collection of $K$ games by $\{G_1,\ldots,G_K\}$, with objective $o_k$ for game $G_k$. 
We train each game separately with population-based self-play and use the late-window policy pairs to build the detector curriculum. 
We use simulated play among learned policies to characterize the family and pick robust training opponents~\citep{wellman2006methods,tuyls2018generalised,wellman2024egta}. 
For the game rewards $\tilde{R}_{\text{eng}}$, the engagement score $R_{\text{eng}}$ standardized on BC rollouts from the training split and clipped to $[-1, 1]$.

\textbf{Detector reward.}
The detector pays a per-step penalty $c_{\text{wait}}>0$ for waiting and receives a classification reward at step $T$:
\begin{equation}
r_{d,t} =
\begin{cases}
- c_{\text{wait}} & t < T, \\[2pt]
R_{\text{cls}}(a^d_T, y) & t = T,
\end{cases}
\label{eq:detector_reward}
\end{equation}
\begin{equation*}
R_{\text{cls}}(a^d_T, y) = \begin{cases}
+1 & (a^d_T,y)\in\{(\textsc{flag},1),(\textsc{pass},0)\}, \\
-1 & (a^d_T,y)=(\textsc{flag},0), \\
-2 & (a^d_T,y)=(\textsc{pass},1).
\end{cases}
\end{equation*}
The larger penalty for missed IO encodes the operational cost asymmetry, and the wait cost rewards early detection. 

\input{figs/algorithm_v2}
\textbf{Real (non-adaptive control).}
The agent is fixed to the cloned policy, $\pi_g \equiv \pi_g^{\text{BC}}$, so its late-window policy is $\bar{\pi}_g^k = \pi_g^{\text{BC}}$ and only the detector trains. 

\textbf{Evade (terminal reward).}
\begin{equation}
r_{g,t}^{\text{Evade}} = \begin{cases}
+1 & t = T,\; a^d_T = \textsc{pass} \\
-1 & t = T,\; a^d_T = \textsc{flag} \\
0 & t < T.
\end{cases}
\label{eq:evade}
\end{equation}

\textbf{Engage (dense reward).}
The agent acts only after \textsc{wait}, so it chooses actions at steps $t_0,\ldots,T-1$, and the dense engagement reward scores the post rendered from each of them,
\begin{equation}
r_{g,t}^{\text{Engage}} = \tilde{R}_{\text{eng}}(s_t, a_t),
\qquad t_0 \le t < T.
\label{eq:engage}
\end{equation}

\textbf{Evade+Engage (E+E).}
A convex combination of Equations~\eqref{eq:evade}--\eqref{eq:engage} would mix a one-time reward with a per-step one, so longer episodes would weight engagement more just by having more steps. We instead average engagement over the episode and deliver the combination as one terminal signal,
\begin{equation}
r_{g,T}^{\text{E+E}} = \alpha\,r_{g,T}^{\text{Evade}} + \frac{1-\alpha}{T-t_0}\sum_{t=t_0}^{T-1} r_{g,t}^{\text{Engage}},
\label{eq:joint_reward}
\end{equation}
We set $\alpha=0.5$, with $r_{g,t}^{\text{E+E}}=0$ for $t<T$. The engagement term is omitted when $T=t_0$.

\subsection{Co-evolutionary Training Procedure}
\label{sec:training}
Each adaptive game is a two-player general-sum game. 
Under Evade the agent's terminal reward is $\pm1$ while the detector's is $+1$ or $-2$, the detector alone pays the wait cost, and control episodes reward only the detector. 
Let $o$ be the game's objective and $\nu^{\pi_g,\pi_d}$ the distribution of states visited when $\pi_g$ plays $\pi_d$. 
The agent maximizes
\begin{equation}
\begin{aligned}
J_g(\pi_g;\pi_d) ={}& \mathbb{E}_{\pi_g,\pi_d}\Big[\sum_{t=t_0}^{T}\gamma^{\,t-t_0}\,r^{o}_{g,t}\Big] \\
&-\lambda_{\text{KL}}\,\mathbb{E}_{s\sim \nu^{\pi_g,\pi_d}}\Big[D_{\mathrm{KL}}\big(\pi_g(\cdot\mid s)\,\|\,\pi_g^{\text{BC}}(\cdot\mid s)\big)\Big],
\end{aligned}
\label{eq:agent_objective}
\end{equation}
with $\lambda_{\text{KL}}>0$ in the BC-initialized games and $\lambda_{\text{KL}}=0$ in the EM and SD games. The expectation runs over IO episodes only, because the agent does not act in control episodes. The detector maximizes
\begin{equation}
J_d(\pi_d;\pi_g)=\mathbb{E}_{\pi_g,\pi_d,\,\mathcal{D}_{\text{ctrl}}}\Big[\sum_{t=t_0}^{T}\gamma^{\,t-t_0}\,r_{d,t}\Big],
\label{eq:detector_objective}
\end{equation}
where the expectation runs over IO episodes generated by $\pi_g$ and control episodes replayed from $\mathcal{D}_{\text{ctrl}}$. 

\textbf{Learning rule.}
Both players are trained with PPO~\citep{schulman2017proximal} on an alternating schedule, as in adversarial reinforcement learning~\citep{pinto2017robust}: each iteration collects an IO batch, updates the agent for $n_{\text{ep}}$ epochs with the detector fixed, then updates the detector on the same or a stored batch plus control episodes with the agent fixed. 
In BC--Real only the detector learns. 
Each player receives the same number of co-evolution updates per run, the detector's $N_w$ warm-up updates aside, so each update is a partial best response and neither gains a budget advantage 
(More details in Appendix~\ref{sec:ppo_app}).

\textbf{Opponent and batch sampling.}
Training only against the current opponent lets each player overfit to a moving target and forget strategies the opponent no longer plays. 
We therefore keep every detector snapshot and every agent batch: with probability $\rho$ the agent plays a snapshot drawn uniformly from the history and the detector trains on a batch drawn uniformly from the history; otherwise each uses the current one (Algorithm~\ref{alg:ioai}, lines~\ref{ln:opp} and~\ref{ln:detppo}). 
This is the opponent model of fictitious self-play~\citep{heinrich2015fictitious,heinrich2016deep} and of \citet{bansal2018emergent}; unlike PSRO and league training~\citep{lanctot2017unified,vinyals2019alphastar}, it uses no meta-solver and no exploiter agents (Appendix~\ref{sec:replay_app}). 

\textbf{Schedule and outcome.}
In each game the detector starts from random weights and first takes $N_w$ warm-up updates against the frozen initial agent; co-evolution then runs for at most $N_a$ iterations or until the window-mean returns and validation MissRate are stable across two consecutive windows of $W$ iterations (Appendix~\ref{sec:schedule_app}). 
The outcome of game $G_k$ is the late-window pair $(\bar{\pi}_g^k,\bar{\pi}_d^k)$, uniform mixtures over each player's final $W$ snapshots (line~\ref{ln:avg}). 
Algorithm~\ref{alg:ioai} gives the full pipeline, from initialization (lines~\ref{ln:bc}--\ref{ln:sd}) through game solving (\textsc{SolveGame}, Phase~1) to curriculum training (\textsc{Curriculum}, Phase~2).

\subsection{Curriculum Training}
\label{sec:curriculum}
Agent rewards are not comparable across objectives, so we use a common difficulty measure for the easy-to-hard curriculum~\citep{bengio2009curriculum}.

\textbf{Difficulty metric.}
We compute the cross-play matrix $M_{kk'}$, the MissRate of agent $\bar{\pi}_g^k$ against detector $\bar{\pi}_d^{k'}$, and average each agent's row across all $K$ late-window detectors, including its own, to obtain its difficulty $\bar{M}_k$ (Algorithm~\ref{alg:ioai}, lines~\ref{ln:crossplay}--\ref{ln:sort}). Higher difficulty indicates greater survival across the learned detector population. Cross-play, sampling and pacing use a curriculum-validation split of campaign contexts and prefixes disjoint from gradient-training rollouts and final evaluation campaigns.

\textbf{Curriculum construction.}
We sort the frozen agents by increasing difficulty, relabel them so that $\bar{M}_1 \le \cdots \le \bar{M}_K$, and initialize the active set $\mathcal{C}$ with the easiest agent (lines~\ref{ln:sort} and~\ref{ln:initc}). With $|\mathcal{C}|$ active agents, we sample a game index according to
\begin{equation}
p_k=\frac{\exp(\lambda_{\text{c}}\hat{M}_k)}{\sum_{k'=1}^{|\mathcal{C}|}\exp(\lambda_{\text{c}}\hat{M}_{k'})},
\qquad k\in\{1,\ldots,|\mathcal{C}|\},
\label{eq:curriculum_sampling}
\end{equation}
where $\hat{M}_k$ is the curriculum-validation estimate of agent $k$'s MissRate against the current curriculum detector and $\lambda_{\text{c}}\ge0$ controls concentration (line~\ref{ln:sampler}). Only the curriculum detector is updated with PPO on sampled IO and matched-control trajectories (line~\ref{ln:currppo}). We add the next agent when both the active-set MissRate and matched-control FalseAlarm meet thresholds $\eta_m$ and $\eta_f$, or when the maximum dwell time $N_c$ is reached (line~\ref{ln:gate}). Once all $K$ agents are active, training continues with the same sampling until the budget of $N_{\text{curr}}$ detector updates is spent.

%% file: figs/fig_framework.tex
\begin{figure*}[t]
\centering
\includegraphics[width=\linewidth]{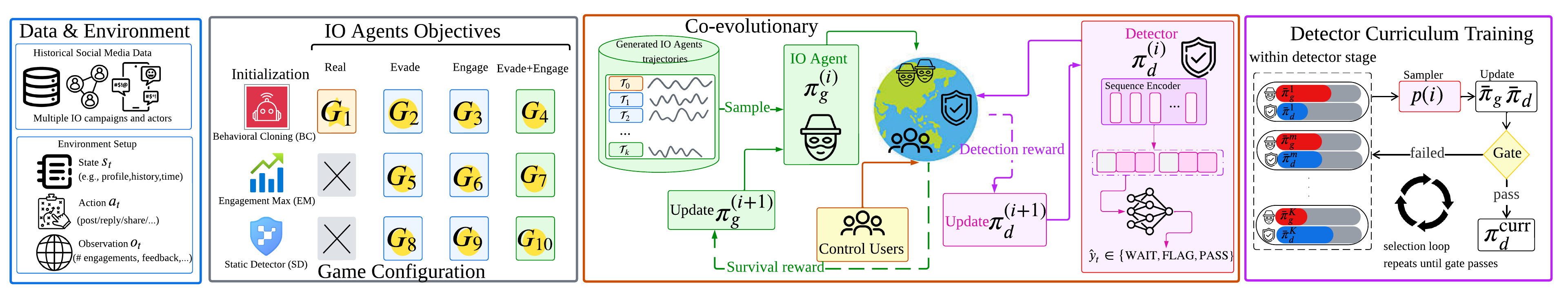}
\caption{Overview of the IO-AI framework for information operation detection through agent--detector co-evolution and adversarial curriculum learning.}
\label{fig:framework}
\end{figure*}

%% file: figs/algorithm_v2.tex
\begin{algorithm}[!t]
\caption{IO-AI framework}
\label{alg:ioai}
\small
\renewcommand{\algorithmicrequire}{\textbf{Input:}}
\renewcommand{\algorithmicensure}{\textbf{Output:}}
\begin{algorithmic}[1]
\Require IO archives $\mathcal{D}_{\text{IO}}$, controls $\mathcal{D}_{\text{ctrl}}$, renderer $q_{\text{txt}}$, engagement estimator $R_{\text{eng}}$, reference detector $\pi_d^{\text{ref}}$, games $G_1,\dots,G_K$
\Ensure curriculum detector $\pi_d^{\text{curr}}$
\State $\pi_g^{\text{BC}} \gets \arg\min_{\pi_g}\mathcal{L}_{\text{BC}}$ on $\mathcal{D}_{\text{IO}}$\Comment{initial policies}\label{ln:bc}
\State $\pi_g^{\text{EM}} \gets \mathrm{DPO}(\pi_g^{\text{BC}};\,\mathcal{P}_{\text{eng}})$, pairs scored by $R_{\text{eng}}$ with margin $\delta$\label{ln:em}
\State $\pi_g^{\text{SD}} \gets \mathrm{PPO}(\pi_g^{\text{BC}};\,r_g^{\text{Evade}})$ against the frozen $\pi_d^{\text{ref}}$\label{ln:sd}
\For{$k=1,\dots,K$}\Comment{Phase 1, solve each game}
  \State $(\bar{\pi}_g^k,\bar{\pi}_d^k) \gets \Call{SolveGame}{G_k}$
\EndFor
\State $M_{kk'} \gets \mathrm{MissRate}(\bar{\pi}_g^k,\bar{\pi}_d^{k'})$ for all $k,k'$\Comment{Phase 2, cross-play difficulty}\label{ln:crossplay}
\State $\bar{M}_k \gets K^{-1}\sum_{k'}M_{kk'}$;\; relabel games so that $\bar{M}_1\le\cdots\le\bar{M}_K$\label{ln:sort}
\State \Return \Call{Curriculum}{$\bar{\pi}_g^1,\dots,\bar{\pi}_g^K$}
\Statex
\Function{SolveGame}{$G_k$ with initialization $\pi_g^{(0)}$ and objective $o_k$}
  \State init $\pi_d^{(0)}$ at random
  \State update $\pi_d^{(0)}$ for $N_w$ warm-up iterations against $\pi_g^{(0)}$\label{ln:warmup}
  \For{$i=0,\dots,N_a-1$}
    \State $\pi_d^{\text{opp}} \gets$ uniform draw from $\{\pi_d^{(j)}\}_{j\le i}$ w.p.\ $\rho$, else $\pi_d^{(i)}$\label{ln:opp}
    \State $\mathcal{T}_i \gets$ IO episodes of $\pi_g^{(i)}$ against $\pi_d^{\text{opp}}$
    \State $\pi_g^{(i+1)} \gets$ PPO update on $\mathcal{T}_i$ for $J_g$ (Eq.~\ref{eq:agent_objective})\Comment{skipped if $o_k=\text{Real}$}\label{ln:agentppo}
    \State $\mathcal{T}_d \gets$ uniform draw from $\{\mathcal{T}_j\}_{j\le i}$ w.p.\ $\rho$, else $\mathcal{T}_i$
    \State run $\pi_d^{(i)}$ on IO histories in $\mathcal{T}_d$ and controls from $\mathcal{D}_{\text{ctrl}}$ to collect detector episodes
    \State $\pi_d^{(i+1)} \gets$ PPO update on these episodes for $J_d$ (Eq.~\ref{eq:detector_objective})\label{ln:detppo}
    \If{$i+1\ge2W$ and window-mean returns and validation MissRate are stable over two consecutive $W$-iteration windows}\label{ln:break}
      \State \textbf{break}
    \EndIf
  \EndFor
  \State \Return uniform mixtures of each player's last $W$ snapshots\label{ln:avg}
\EndFunction
\Statex
\Function{Curriculum}{$\bar{\pi}_g^1,\dots,\bar{\pi}_g^K$}
  \State init $\pi_d^{\text{curr}}$;\; $\mathcal{C}\gets\{\bar{\pi}_g^1\}$;\; $n_c\gets0$\Comment{easiest agent first}\label{ln:initc}
  \For{$i=1,\dots,N_{\text{curr}}$}
    \State $\hat{M}_k \gets \widehat{\mathrm{MissRate}}(\bar{\pi}_g^k,\pi_d^{\text{curr}})$ for each $\bar{\pi}_g^k\in\mathcal{C}$\label{ln:active}
    \State sample $k$ with $p_k\propto\exp(\lambda_{\text{c}}\hat{M}_k)$ (Eq.~\ref{eq:curriculum_sampling})\Comment{hardest sampled most}\label{ln:sampler}
    \State $\pi_d^{\text{curr}} \gets$ PPO step on $\bar{\pi}_g^k$ and $\mathcal{D}_{\text{ctrl}}$ episodes;\; $n_c\gets n_c+1$\label{ln:currppo}
    \If{$n_c\ge N_c$ or ($\widehat{\mathrm{MissRate}}(\mathcal{C})\le\eta_m$ and $\widehat{\mathrm{FalseAlarm}}\le\eta_f$)}\label{ln:gate}
      \State add $\bar{\pi}_g^{|\mathcal{C}|+1}$ to $\mathcal{C}$ if $|\mathcal{C}|<K$;\; $n_c\gets0$\Comment{expand the pool}
    \EndIf
  \EndFor
  \State \Return $\pi_d^{\text{curr}}$
\EndFunction
\end{algorithmic}
\end{algorithm}

%% file: sections/exp_v5.tex
\input{figs/tab_dataset}
\input{figs/tab_main_results}
\section{Experiments}
\label{sec:experiments}
We ask how engagement and evasion objectives change the behavior a detector must recognize, whether training across these behaviors improves coverage, and how far that coverage extends across actors and periods.

\subsection{Datasets}
\label{subsec:dataset}
We use $20$ IO campaigns from Twitter's archive of state-backed information operations\footnote{\url{https://blog.x.com/en_us/topics/company/2021/-expanding-access-beyond-information-operations-}} (October 2018--February 2021), covering eight state-actor categories (Table~\ref{tab:dataset}). 
Twelve campaigns are used for behavioral cloning, engagement-model training and detector training; eight are held out to evaluate later campaigns from known actors and campaigns from unseen actors.
\footnote{All training stages exclude the held-out campaigns and their evaluation periods (Appendix~\ref{sec:data_eval_app}); We retain campaigns with at least $60\%$ English posts to match the pretrained text generator.} 
Real non-IO histories provide controls matched by topic and activity volume; evaluation controls are matched to held-out campaigns.

\subsection{Baselines and Evaluation}
\label{sec:eval}
We compare with six state-of-the-art baselines. 
Four use per-account information: \textbf{RF}, a random forest over behavioral aggregates~\citep{varol2017online}; 
\textbf{BERT}, fine-tuned on concatenated account posts~\citep{devlin2019bert}; \textbf{Propagation RNN}, a recurrent and convolutional classifier over propagation paths~\citep{liu2018early}; and \textbf{MetaTroll}, a few-shot troll detector with campaign-specific adapters~\citep{tian2023metatroll}. 
Two use cross-account structure: the \textbf{fused coordination graph}, which combines behavioral-similarity networks~\citep{luceri2024unmasking}, and \textbf{IOHunter}, which combines network and language-model representations~\citep{minici2025iohunter}. 
These baselines classify completed trajectories; we report their $F_1$ in Table~\ref{tab:main_results}.
We also evaluate four variants of our framework: \textbf{single-campaign static} and \textbf{multi-campaign static}, trained with supervision on one and all 12 campaigns, respectively; \textbf{single-game adversarial}, trained from random weights by co-evolution with BC-, EM-, or SD-initialized agents; and \textbf{curriculum}, trained against agents from all 10 games in easy-to-hard order. 

\textbf{Evaluation metrics.}
We report macro $F_1$ on balanced test sets. 
For sequential detectors, MissRate and FalseAlarm measure terminal errors (Equation~\ref{eq:terminal_error_rates}); Delay measures mean steps to flag among detected IO accounts. 
Survival@$H$ is a secondary Kaplan--Meier diagnostic that censors \textsc{pass} at termination~\citep{kaplan1958nonparametric}. 
\footnote{Because censoring depends on the detector's decision, this diagnostic is not an unbiased evasion probability under continued observation (Appendix~\ref{sec:survival_estimation}).}
Appendix~\ref{sec:ablations} checks three design choices: the autoregressive policy, the BC anchor and the renderer. 

\input{figs/fig_tradeoffs}

\subsection{Detector Performance Across Tests}
\label{sec:results_detectors}
A deployed detector meets operations it did not train on. 
Even content features that separate an archived campaign from organic activity are ``quite dynamic'', transfer poorly between campaigns of one actor, and cannot be expected to catch ``activity that is not yet known''~\citep{alizadeh2020content}. 
Table~\ref{tab:main_results} reports each detector where the actor, the period or the strategy changes, alongside its in-domain score. 

In-domain, all baselines sit within $9$ pp of one another ($0.83$--$0.92$). 
Under shift the spread opens to $35$ pp on an unseen actor's later period and $39$ pp on Adaptive-SD, and in-domain rank is a weak guide to it: text-only BERT beats the propagation-path RNN in-domain ($0.88$ versus $0.86$) and scores below it in every held-out condition.
The per-account detectors transfer worst. BERT falls from $0.88{\pm}0.01$ in-domain to $0.52{\pm}0.05$ on an unseen actor and $0.44{\pm}0.06$ on that actor's later period, and to $0.41{\pm}0.06$ and $0.26{\pm}0.05$ against the BC and SD agents, within $2$ pp of chance or below it in four of the six held-out conditions; it keeps a margin over chance only on the known actor's later period ($0.66{\pm}0.04$) and against EM agents ($0.59{\pm}0.04$). The account-feature RF is lower still on each of the four ($0.44$, $0.37$, $0.31$, $0.21$). The propagation RNN carries across time better than across actors: it keeps $0.69{\pm}0.04$ on the known actor's later period, $17$ pp below in-domain, and $0.57{\pm}0.05$ on an unseen actor, and it falls to $0.43{\pm}0.05$ and $0.32{\pm}0.05$ against the BC and SD agents. MetaTroll's campaign adapters give the best cross-actor transfer among the four ($0.65{\pm}0.04$ on an unseen actor) and reach $0.51{\pm}0.05$ and $0.39{\pm}0.05$ against the same two agents.
The fused coordination graph and IOHunter hold $0.70{\pm}0.04$ and $0.79{\pm}0.03$ on an unseen actor and $0.63{\pm}0.05$ and $0.72{\pm}0.04$ on its later period, $5$--$16$ pp above the best per-account detector. 
Interestingly, Adaptive-EM is the easiest of the three adaptive conditions for all six published detectors and Adaptive-SD the hardest for five of them.

IO-AI has the highest mean $F_1$ in all six shifted or adaptive conditions, and its in-domain score is $0.91$ against $0.92$ for IOHunter. Its margin also holds against fresh adaptation: when a new Evade agent is trained against each frozen detector for $100$ PPO iterations, full IO-AI keeps the lowest final miss rate for every initialization, $0.18$ to $0.23$ (Appendix~\ref{sec:reattack_app}).

\subsection{How Do Objectives Change Behavior?}
\label{sec:results_games}
Engagement and evasion place different pressures on an agent's posting behavior. 
We measure both pressures on the nine adaptive configurations against the fixed BC--Real control (Figure~\ref{fig:tradeoffs}a).
Survival gain, $\Delta S$, is the change in late-window $\mathrm{Survival@}T_{\max}$, in percentage points, against the same frozen static detector for every configuration. 
Engagement gain, $\Delta E$, is the change in mean predicted engagement per post.

\textbf{The same evasion objective produces different behavioral policies across starting configurations.}
Claim-makers draw on an inherited repertoire and innovate at its margins~\citep{tilly2015contentious}; if the same holds for an operation, its posting habits should constrain how it adapts. 
The three Evade agents receive the same reward and reach it by different policies. 
SD--Evade waits longer between actions and retweets less than BC--Real, and both measures move toward control users (Figure~\ref{fig:case_study_retweet}). 
It sits among controls in the static detector's pass region (Figure~\ref{fig:limits_static}), and the detector takes $9.1$ steps on average to decide on it, against $3.2$ for BC--Real. 
EM--Evade keeps the displacement from the BC anchor that engagement pretraining gave it (Appendix Figure~\ref{fig:training_dynamics}), together with an engagement gain near $+19$ reactions per post, where BC--Evade and SD--Evade stay near $+3$ (Figure~\ref{fig:tradeoffs}a). 
BC--Evade stays closest to the anchor throughout training. The survival gains differ accordingly: $+32$, $+25$ and $+17$ pp for SD, EM and BC. 
Across all 45 adaptive runs, objective, initialization and their interaction account for $69\%$, $18\%$ and $8\%$ of the variance in late-window survival gain. 

\input{figs/fig_case_study_summary}
\input{figs/tab_operating_metrics_main}
\textbf{Under the same engagement objective, only the EM start turns its behavioral shift into engagement.}
EM--Engage gains approximately $45$ engagement-score units and $2$ pp in survival. 
BC--Engage and SD--Engage gain $6$ and $5$ units and lose $11$ and $5$ pp. 
The start that arrives with engagement pretraining raises its score seven to nine times more than the two that must learn it in the game, and it is the only one that pays no survival for the gain. 
Figure~\ref{fig:case_study_emotion} shows the behavioral policy EM--Engage learns from the reward: retweets at near one-minute intervals with emotion intensity of $0.7$ to $0.9$, against about ten minutes and $0.3$ to $0.5$ for BC--Real.
\citet{brady2017emotion} also found that each moral-emotional word in a post raises its retweet rate by about $20\%$ across $563{,}312$ tweets on three political issues, and high-arousal content travels further in general~\citep{berger2012what}. 
Interestingly, the same rapid and emotional retweeting is also the pattern the archived campaigns were caught producing and the static detector learned to flag. 
An agent that acquires it becomes easier to recognize. Across engage-only training, the frozen detector's performance rises from $0.73$ to $0.92$.

The dual-objective games test whether agents can raise predicted engagement while keeping their evasion gains (Figure~\ref{fig:tradeoffs}a). 
Relative to Evade, BC and EM each lose $12$ pp of survival gain and gain about $5$ and $21$ engagement scores. 
SD shows a weaker trade-off: it gained approximately $11$ engagement-score units while losing only $3$ pp of survival gain, from $32$ to $29$ pp. 
We also find that the EM and SD outcomes sit on different parts of the observed frontier. 
EM reaches higher predicted engagement and SD keeps more survival gain; neither dominates the other, and BC's joint outcome is below both on each axis. 
Goal-systems theory says when that is possible: goals conflict when they compete for the same means~\citep{kruglanski2002theory}, and serving several goals at once narrows the means that remain acceptable~\citep{kopetz2011multifinality}.

\subsection{How Well Do Detectors Transfer?}
\label{sec:results_transfer}
Section~\ref{sec:results_detectors} showed how far the published detectors fall under shift. 
Here we ask the same of our own: how much of IO-AI's accuracy survives a new actor or a later period, and where it comes from. 
Figure~\ref{fig:actor_dotplot} scores each actor group on its held-out campaigns, and Table~\ref{tab:time_split} trains on one period and tests on a later one.

\textbf{Across actors.}
On unseen actors the full IO-AI detector reaches $F_1=0.81{\pm}0.03$, against $0.62{\pm}0.05$ for static training on the same twelve campaigns and $0.79{\pm}0.03$ for IOHunter (Table~\ref{tab:main_results}). 
From its in-domain score it drops $10$ pp; static training drops $28$. 
IO-AI is above static training in all eleven actor groups, by $13$ to $24$ pp (Figure~\ref{fig:actor_dotplot}), and the gains run opposite to representation: smallest for Russia/IRA ($13$ pp) and Iran ($15$ pp), the actors with the most campaigns in the training pool, and largest for Ghana/Nigeria ($24$ pp), which has none. 
The same ordering holds at a fixed operating point. 
With thresholds set to a $10\%$ false-positive rate, recall rises by $4$ pp on IRA 2018 and $3$ pp on China 2019, which static training already handles, and from $0.42$ to $0.73$ on GRU, which it does not (Table~\ref{tab:op_recall}). 
A qualitative read of the GRU accounts in Twitter's February 2021 release~\citep{twitter2021feb} shows what static training was missing: $87\%$ of their tweets are posted through Twitter's web client or web app rather than a scheduling tool, only $14\%$ are retweets (Table~\ref{tab:archives}), and their most-liked posts are original dispatches (Appendix~\ref{sec:archives_engagement_app}).

\textbf{Across periods.}
If a detector's accuracy on a later period came from the period just before it, withholding that period should cost the most. 
As expected, every detector scores lower when the intervening year is missing. 
On the three splits that train through the preceding period, IO-AI reaches $0.83$, $0.88$ and $0.86$, against $0.69$, $0.78$ and $0.74$ for static training and $0.74$, $0.81$ and $0.78$ for the best single-game detector (Table~\ref{tab:time_split}). 
On the split that trains on 2018--2019 and tests on 2021 with 2020 withheld, static training falls from $0.74$ to $0.55$, the best single-game detector from $0.78$ to $0.66$, and IO-AI from $0.86$ to $0.79$: drops of $19$, $12$ and $7$ pp. 
IO-AI's margin over static training is largest, at $24$ pp, where the preceding period is absent. 
The balanced conditions in Table~\ref{tab:main_results} agree with $0.86{\pm}0.02$ against $0.76{\pm}0.03$ on a known actor's later period, and $0.74{\pm}0.03$ against $0.50{\pm}0.05$ on an unseen actor's later period, the hardest condition for every detector in the table. 
This is consistent with interaction patterns that persist across platforms and decades while content and platforms change~\citep{avalle2024persistent}.

%% file: figs/tab_dataset.tex
\begin{table}[t]
\centering
\scriptsize
\setlength{\tabcolsep}{2.5pt}
\caption{IO campaign data statistics.}
\label{tab:dataset}
\resizebox{\columnwidth}{!}{%
\begin{tabular}{lcrrr}
\toprule
Campaign & Time & Users & \# Posts & \# RTs \\
\midrule
Iran 10/18         & Oct 2018 & 770    & 1,122,936  & 232,337    \\
IRA 10/18          & Oct 2018 & 3,608  & 8,768,633  & 3,332,424  \\
Venezuela 1/19     & Jan 2019 & 1,951  & 9,935,542  & 2,054,749  \\
Iran 1/19          & Jan 2019 & 2,311  & 4,671,950  & 2,500,811  \\
Russia 1/19        & Jan 2019 & 416    & 920,761    & 712,800    \\
Ecuador 4/19       & Apr 2019 & 1,019  & 700,240    & 577,749    \\
Iran 6/19          & Jun 2019 & 3,602  & 3,519,943  & 2,202,263  \\
China 7/19         & Jul 2019 & 5,241  & 13,847,731 & 3,169,578  \\
Honduras 2/20      & Feb 2020 & 3,104  & 1,165,019  & 611,717    \\
Egypt 2/20         & Feb 2020 & 2,541  & 7,935,329  & 4,067,645  \\
Indonesia 2/20     & Feb 2020 & 795    & 2,700,296  & 401,803    \\
Serbia 2/20        & Feb 2020 & 8,558  & 40,357,708 & 34,523,930 \\
\midrule
\multicolumn{5}{l}{\emph{Held-out external test pool}} \\
Ghana/Nigeria 3/20 & Mar 2020 & 70     & 39,964     & 16,310     \\
China 5/20         & May 2020 & 23,750 & 348,608    & 182,002    \\
Iran 9/20          & Sep 2020 & 104    & 2,450      & 551        \\
Russia 9/20        & Sep 2020 & 1,158  & 3,436,160  & 416,435    \\
Saudi Arabia 9/20  & Sep 2020 & 33     & 220,254    & 163,470    \\
Iran 2/21          & Feb 2021 & 238    & 560,571    & 100,446    \\
Russia IRA 2/21    & Feb 2021 & 31     & 68,914     & 21,775     \\
Russia GRU 2/21    & Feb 2021 & 70     & 26,684     & 3,746      \\
\bottomrule
\end{tabular}}
\end{table}

%% file: figs/tab_main_results.tex
\begin{table*}[t]
\centering
\caption{Detector macro $F_1$ on balanced test sets (mean $\pm$ standard deviation over five seeds; higher is better).}
\label{tab:main_results}
\resizebox{\textwidth}{!}{%
\begin{tabular}{lccccccc}
\toprule
Detector & \makecell{In-\\domain} & \makecell{Known-actor\\future} & \makecell{Unseen\\actor} & \makecell{Unseen-actor\\future} & \makecell{Adaptive\\BC} & \makecell{Adaptive\\EM} & \makecell{Adaptive\\SD} \\
\midrule
RF~\citep{varol2017online} & $0.83{\pm}0.017$
& $0.57{\pm}0.044$
& $0.44{\pm}0.056$
& $0.37{\pm}0.060$
& $0.31{\pm}0.064$
& $0.49{\pm}0.048$
& $0.21{\pm}0.050$ \\
BERT~\citep{devlin2019bert} & $0.88{\pm}0.013$
& $0.66{\pm}0.039$
& $0.52{\pm}0.050$
& $0.44{\pm}0.055$
& $0.41{\pm}0.059$
& $0.59{\pm}0.043$ & $0.26{\pm}0.051$ \\
Propagation RNN~\citep{liu2018early} & $0.86{\pm}0.016$
& $0.69{\pm}0.036$
& $0.57{\pm}0.046$
& $0.48{\pm}0.050$
& $0.43{\pm}0.054$
& $0.62{\pm}0.041$ & $0.32{\pm}0.053$ \\
MetaTroll~\citep{tian2023metatroll} & $0.89{\pm}0.012$
& $0.75{\pm}0.031$
& $0.65{\pm}0.040$
& $0.56{\pm}0.046$
& $0.51{\pm}0.050$
& $0.69{\pm}0.037$ & $0.39{\pm}0.054$ \\
Fused graph~\citep{luceri2024unmasking} & $0.89{\pm}0.015$ & $0.77{\pm}0.033$ & $0.70{\pm}0.040$ & $0.63{\pm}0.046$ & $0.54{\pm}0.052$ & $0.73{\pm}0.034$ & $0.59{\pm}0.048$ \\
IOHunter~\citep{minici2025iohunter} & $\mathbf{0.92{\pm}0.010}$ & $0.83{\pm}0.024$ & $0.79{\pm}0.030$ & $0.72{\pm}0.036$ & $0.63{\pm}0.043$ & $0.79{\pm}0.029$ & $0.60{\pm}0.047$ \\
\midrule
\multicolumn{8}{@{}l}{\emph{IO-AI (ours)}} \\
Static, 1 campaign & $0.91{\pm}0.008$ & $0.61{\pm}0.047$ & $0.43{\pm}0.061$ & $0.42{\pm}0.052$ & $0.36{\pm}0.058$ & $0.53{\pm}0.051$ & $0.19{\pm}0.036$ \\
Static, 12 campaigns & $0.90{\pm}0.014$ & $0.76{\pm}0.034$ & $0.62{\pm}0.046$ & $0.50{\pm}0.049$ & $0.47{\pm}0.055$ & $0.65{\pm}0.044$ & $0.28{\pm}0.043$ \\
Co-evolution, BC agents & $0.88{\pm}0.018$ & $0.74{\pm}0.032$ & $0.64{\pm}0.041$ & $0.55{\pm}0.045$ & $0.75{\pm}0.027$ & $0.67{\pm}0.039$ & $0.31{\pm}0.054$ \\
Co-evolution, EM agents & $0.86{\pm}0.021$ & $0.72{\pm}0.038$ & $0.66{\pm}0.035$ & $0.58{\pm}0.044$ & $0.61{\pm}0.041$ & $0.86{\pm}0.024$ & $0.44{\pm}0.052$ \\
Co-evolution, SD agents & $0.85{\pm}0.024$ & $0.70{\pm}0.042$ & $0.67{\pm}0.033$ & $0.60{\pm}0.039$ & $0.58{\pm}0.049$ & $0.72{\pm}0.036$ & $0.69{\pm}0.031$ \\
\midrule
Population, hard$\to$easy & $0.86{\pm}0.019$ & $0.73{\pm}0.031$ & $0.67{\pm}0.043$ & $0.61{\pm}0.046$ & $0.70{\pm}0.037$ & $0.77{\pm}0.032$ & $0.63{\pm}0.042$ \\
Population, random & $0.89{\pm}0.016$ & $0.80{\pm}0.028$ & $0.74{\pm}0.032$ & $0.68{\pm}0.041$ & $0.76{\pm}0.029$ & $0.82{\pm}0.027$ & $0.70{\pm}0.038$ \\
\textbf{Full IO-AI} (easy$\to$hard) & $0.91{\pm}0.009$ & $\mathbf{0.86{\pm}0.019}$ & $\mathbf{0.81{\pm}0.026}$ & $\mathbf{0.74{\pm}0.031}$ & $\mathbf{0.83{\pm}0.021}$ & $\mathbf{0.87{\pm}0.018}$ & $\mathbf{0.78{\pm}0.028}$ \\
\bottomrule
\end{tabular}}
\end{table*}

%% file: figs/fig_tradeoffs.tex
\begin{figure*}[t]
\centering
\begin{subfigure}[t]{0.285\textwidth}
\centering
\includegraphics[width=\linewidth]{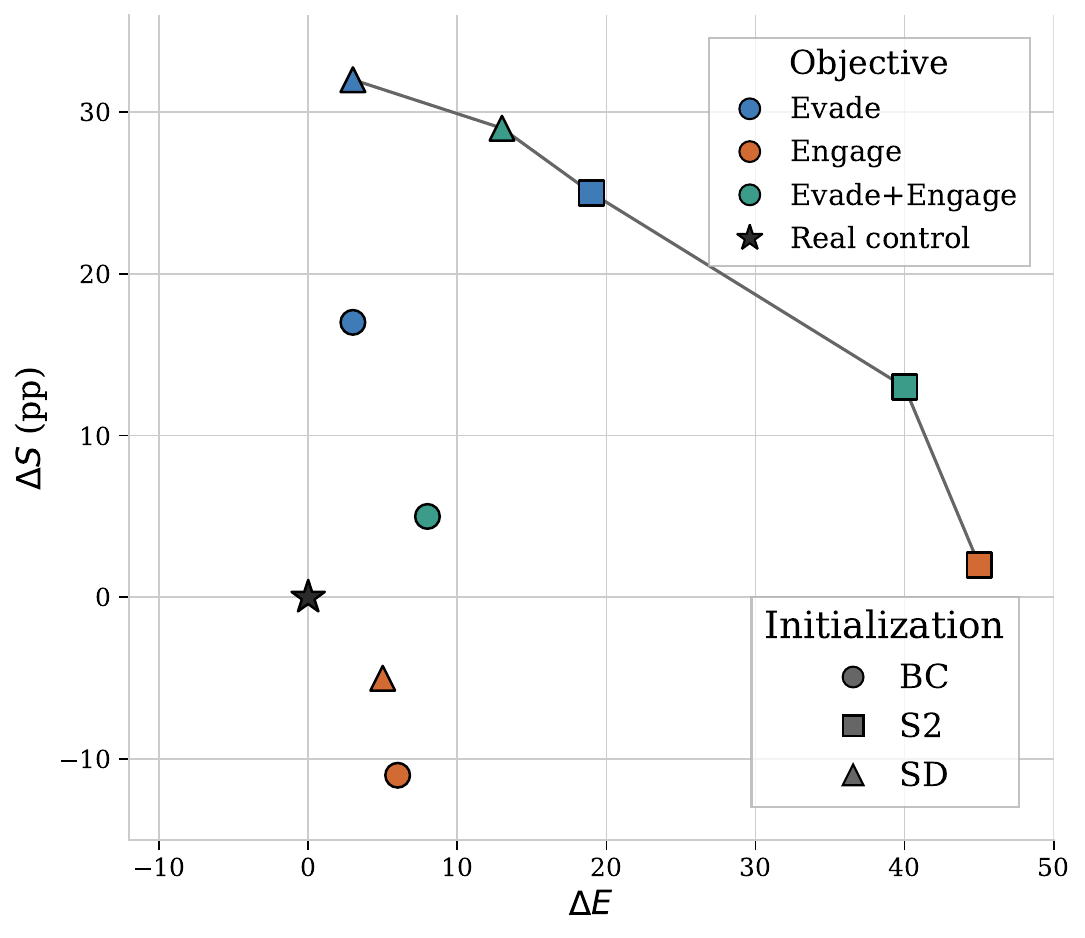}
\caption{Survival--engagement trade-off}
\label{fig:tradeoffs_a}
\end{subfigure}\hfill
\begin{subfigure}[t]{0.345\textwidth}
\centering
\includegraphics[width=\linewidth]{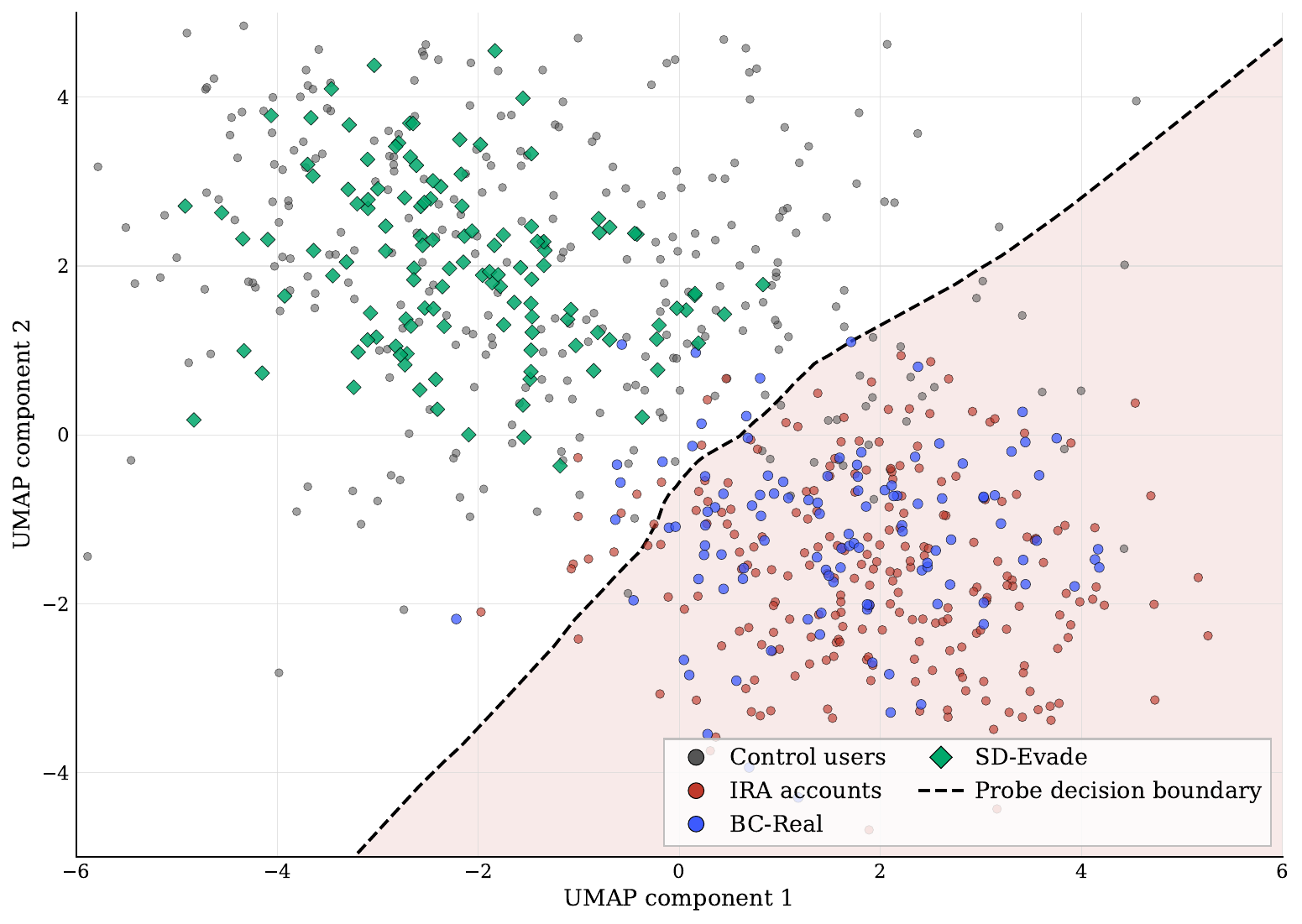}
\caption{Static detector's pass region}
\label{fig:limits_static}
\end{subfigure}\hfill
\begin{subfigure}[t]{0.351\textwidth}
\centering
\includegraphics[width=\linewidth]{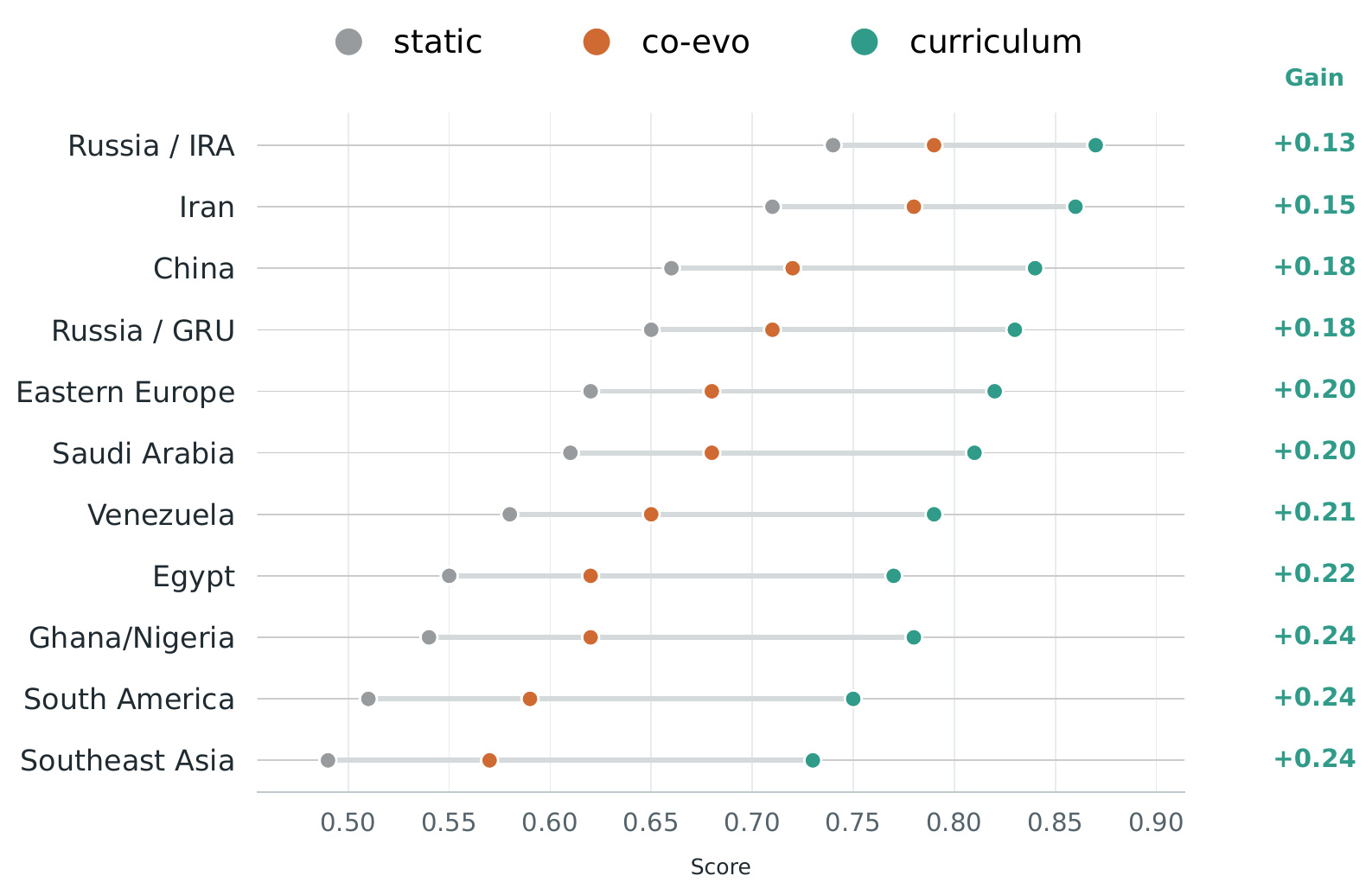}
\caption{Held-out actor gains}
\label{fig:actor_dotplot}
\end{subfigure}
\caption{Agent outcomes and detector coverage.
(a) Late-window survival ($\Delta S$, pp against a common frozen static detector) and predicted engagement ($\Delta E$, reactions per post) gains over BC--Real.
Color denotes objective, shape denotes initialization (S2 = EM), and the gray line shows the empirical frontier across nine games.
(b) UMAP of static-detector representations with a fitted probe boundary; SD--Evade overlaps controls in the pass region.
(c) Held-out actor-group $F_1$ for multi-campaign static, best single-game, and curriculum detectors; gains are curriculum minus static.}
\label{fig:tradeoffs}
\end{figure*}

%% file: figs/fig_case_study_summary.tex
\begin{figure*}[t]
\centering
\begin{subfigure}[b]{0.49\textwidth}
\centering
\includegraphics[width=\linewidth]{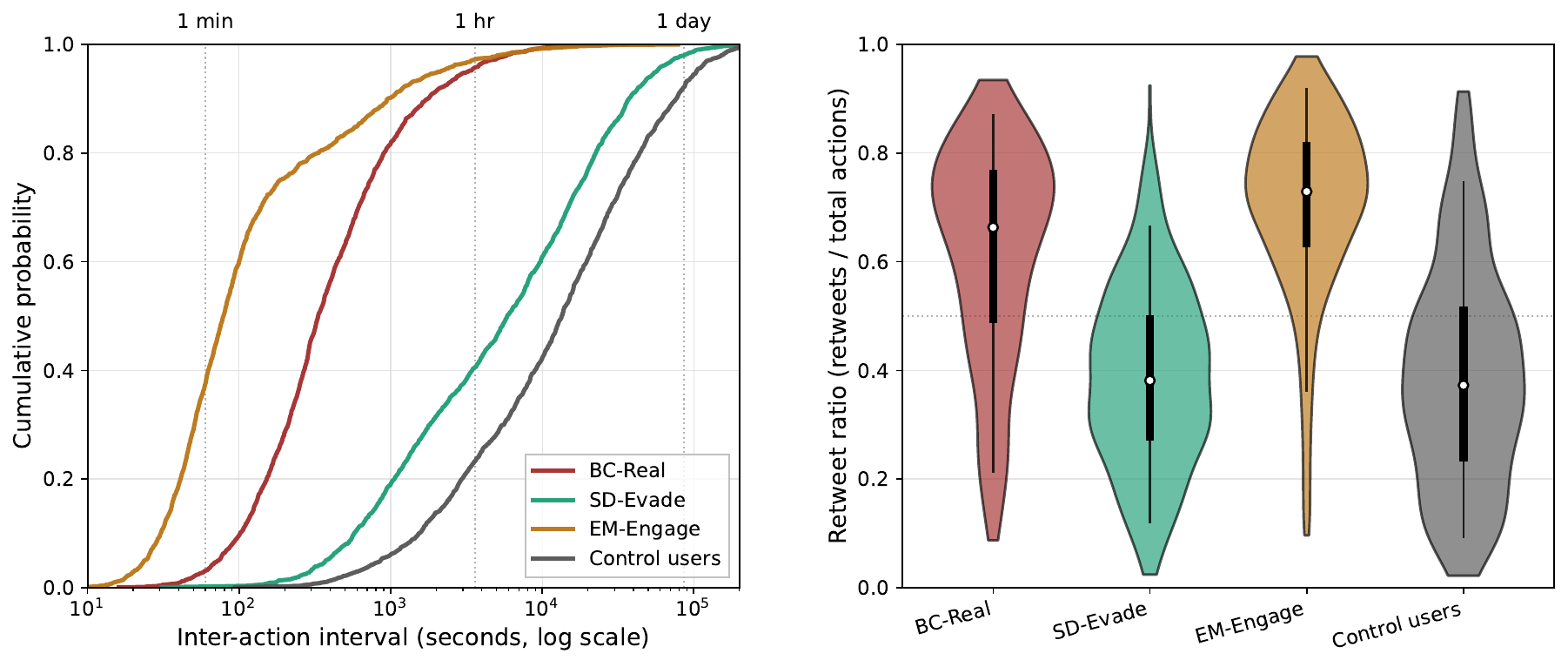}
\caption{Intervals and retweet share}
\label{fig:case_study_retweet}
\end{subfigure}\hfill
\begin{subfigure}[b]{0.49\textwidth}
\centering
\includegraphics[width=\linewidth]{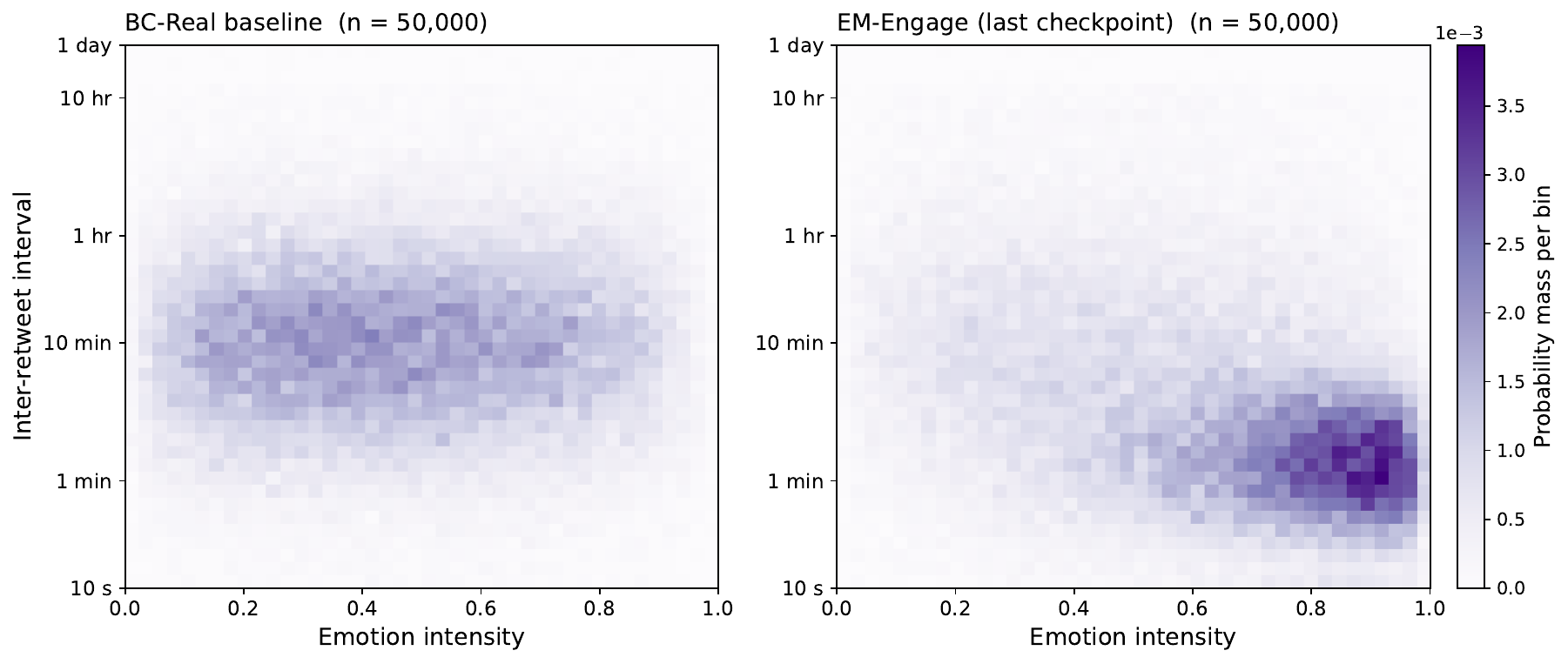}
\caption{Emotion intensity against inter-retweet interval}
\label{fig:case_study_emotion}
\end{subfigure}
\caption{Behavioral profiles of cloned and adapted agents. (a) Inter-action intervals and retweet shares. 
(b) Emotion intensity on a $[0,1]$ scale against inter-retweet interval at the final checkpoint.
}
\label{fig:case_study_summary}
\end{figure*}

%% file: figs/tab_operating_metrics_main.tex
\begin{table}[t]
\centering
\small
\setlength{\tabcolsep}{3pt}
\caption{Sequential operating metrics against adaptive agents (mean $\pm$ SD over five seeds; $\downarrow$ better).}
\label{tab:operating_metrics_main}
\begin{tabular*}{\columnwidth}{@{\extracolsep{\fill}}lcccc}
\toprule
Detector & MissRate & FalseAlarm & Delay & Survival@25 \\
\midrule
Static, 12 campaigns      & $0.43{\pm}0.04$ & $0.12{\pm}0.02$ & $\mathbf{4.8{\pm}0.6}$ & $0.38{\pm}0.05$ \\
Co-evolution, SD agents      & $0.29{\pm}0.04$ & $0.13{\pm}0.02$ & $6.2{\pm}0.7$ & $0.27{\pm}0.04$ \\
Population, random   & $0.22{\pm}0.03$ & $0.11{\pm}0.02$ & $5.8{\pm}0.5$ & $0.20{\pm}0.03$ \\
Full IO-AI & $\mathbf{0.15{\pm}0.02}$ & $\mathbf{0.10{\pm}0.02}$ & $5.1{\pm}0.4$ & $\mathbf{0.14{\pm}0.03}$ \\
\bottomrule
\end{tabular*}
\end{table}

%% file: sections/case_study_v8.tex
\section{Behavioral Distance and Detector Transfer}
\label{sec:results_diagnostic}

The transfer results raise a practical question: which new cohorts need evaluation first?
We examine whether differences between policies cloned from cohort histories are associated with differences in detector transfer. 
For a cohort $A$, defined by actor or time period, let $\pi^A$ be the BC policy (\S\ref{sec:agent}) cloned from its unlabeled trajectories, with heads $\pi^A_e$, $\pi^A_c$, and $\pi^A_\tau$ as in Equation~\eqref{eq:autoregressive}. 
We evaluate all policies on a shared reference batch $\mathcal{B}_{\text{ref}}=\{(s_b,e_b,c_b)\}_{b=1}^{N_{\text{ref}}}$ constructed from held-out trajectories. Each record contains a state and the recorded next action's emotion and posting type. 
With $\mu_{\text{ref}}$ the uniform distribution over the $N_{\text{ref}}$ records, we define
\begin{equation}
d(A,B) = \mathbb{E}_{(s,e,c)\sim\mu_{\text{ref}}}\Big[\sum_{h\in\{e,c,\tau\}} D_h\big(\pi^A_h(\cdot\mid s,\mathbf{a}_{<h}),\,\pi^B_h(\cdot\mid s,\mathbf{a}_{<h})\big)\Big],
\label{eq:policy_distance}
\end{equation}
where $\mathbf{a}_{<h}$ denotes the recorded parent actions of head $h$ (none for $e$, $e$ for $c$, and $(e,c)$ for $\tau$), and $D_h$ is the Jensen--Shannon divergence~\citep{lin1991divergence} for the two categorical heads and the Wasserstein distance $W_1$~\citep{peyre2019computational} for timing, because a shift from minutes to hours should count for more than a reshuffle among nearby intervals. Both timing heads are evaluated on the $\log_{10}$-minute bins of \S\ref{sec:agent}, and $W_1$ uses the distance between bin centers in $\log_{10}$ minutes, so one unit of timing distance is a tenfold change in interval.
The base-2 JS terms lie in $[0,1]$ while the timing depends on the interval representation.
\input{figs/tab_time_split}

We built the distance to find the operation a detector has not seen, and the Chinese cohort shows it doing so, but the case that matters more is the one we did not build it for, an operation that sits nearest the training cohorts and is still missed most.
Greater behavioral distance is associated with lower transfer $F_1$ across the $18$ train--test pairs (Spearman $\rho_s=-0.58$, $n=18$, $p<0.05$), and the distance is computed independently of detector training.
The Chinese cohort is the expected case, with transfer $F_1$ of $0.55$ after IRA training and $0.59$ after GRU training and both policy distances above the median of $0.62$, at $d(\mathrm{CHN},\mathrm{IRA})=0.69$ and $d(\mathrm{CHN},\mathrm{GRU})=0.95$, most of it in posting type and timing~\citep{king2017chinese}.
GRU is not, since it is the cohort nearest the IRA at $d(\mathrm{IRA},\mathrm{GRU})=0.42$ and the one an IRA-trained detector transfers to best ($F_1=0.72$, Figure~\ref{fig:cross_actor}), yet at a $10\%$ false-positive rate the static detector misses $58\%$ of its accounts (Table~\ref{tab:op_recall}).
The cohort the distance places nearest the IRA is the one static training misses most and adaptive training recovers it with $+0.31$ on recall.

%% file: figs/tab_time_split.tex
\begin{table}[t]
\centering
\small
\setlength{\tabcolsep}{3pt}
\caption{Temporal transfer ($F_1$). E20 and L20 denote early and late 2020, and 21* denotes 2021 with 2020 excluded from training. Best single-game is the strongest of the three co-evolution detectors on each split.}
\label{tab:time_split}
\begin{tabular*}{\columnwidth}{@{\extracolsep{\fill}}llccc}
\toprule
Train & Test & Static & Best single-game & Curriculum \\
\midrule
18--19               & E20              & 0.69 & 0.74 & \textbf{0.83} \\
18--E20              & L20              & 0.78 & 0.81 & \textbf{0.88} \\
18--20               & 21               & 0.74 & 0.78 & \textbf{0.86} \\
\midrule
18--19               & 21*              & 0.55 & 0.66 & \textbf{0.79} \\
\bottomrule
\end{tabular*}
\end{table}

%% file: sections/conclusion_v3.tex
\section{Conclusion}
\label{sec:conclusion}

IO-AI trains sequential IO detectors against a population of adaptive behavioral agents ordered by cross-play difficulty. 
It broadens coverage beyond static and single-game training, with an advantage that persists under fresh re-attacks within budget.
The games show why both initialization and objective matter when constructing the opponent population. 
Behavioral distance complements this process by identifying cohort differences associated with weaker transfer. 
Together, the findings support using adaptive opponents to test and improve detector coverage for information operations.

%% file: sections/appendix.tex
\appendix

\section{Data and Evaluation Details}
\label{sec:data_eval_app}

\subsection{Campaign Data and Controls}
Table~\ref{tab:dataset} lists the $20$ campaign releases and which of them are used for training or held out. Behavioral cloning, engagement-model fitting and detector training never see the eight held-out campaigns. 
In-domain evaluation uses users and trajectories from the training campaigns that no training stage saw. Controls are matched by topic and activity volume and follow the same campaign and time splits. The Adaptive-BC, Adaptive-EM and Adaptive-SD columns of Table~\ref{tab:main_results} use new rollouts from the frozen late-window agents, so they test how well a detector covers the opponents the curriculum trained on.

Table~\ref{tab:archives} shows the four complete archive files used in the archival analyses. Their counts refer to these files rather than to the campaign releases in Table~\ref{tab:dataset}, and the comparisons drawn from them are limited to the operations the takedown archives captured.

\input{figs/tab_archives}

\subsection{Engagement in the Archives}
\label{sec:archives_engagement_app}
Table~\ref{tab:archives} also describes the GRU file. There, $87.3\%$ of tweets come from Twitter's web client or web app and $14.0\%$ are retweets. Activity concentrates in 2016--2020, with $15$ earlier tweets. The five most-liked English tweets are all originals, and the most-liked ($157$ likes, November 2019) reports from southern Idlib.

\subsection{Per-Operation Recall Protocol}
\label{sec:op_recall_app}
The static detector trained on 12 campaigns and the full IO-AI detector score every account in each complete archive. Each detector's threshold is set to a $10\%$ false-positive rate on held-out benign controls, and Table~\ref{tab:op_recall} reports account-level recall averaged over five detector seeds. This fixed operating point is the counterpart of the balanced-test-set macro $F_1$ in Table~\ref{tab:main_results}. The within-archive gap between the two detectors is the comparison of interest; recall levels differ across archives for other reasons as well.

\input{figs/tab_op_recall}

\subsection{Survival Estimation}
\label{sec:survival_estimation}
For IO episodes, let $n^{\mathrm{f}}_u$ be the number of flags at decision step $u$ and $n^{\mathrm{r}}_u$ the number of episodes still at risk just before that step. The horizon $H$ and the index $u$ count decision steps (\S\ref{sec:formulation}).
A \textsc{pass} censors the episode at its terminal step and removes it from later risk sets.
The Kaplan--Meier diagnostic of \S\ref{sec:eval} is
\begin{equation*}
\mathrm{Survival@}H=\widehat{S}(H)
=\prod_{\substack{u=1,\ldots,H\\n^{\mathrm{f}}_u>0}}\left(1-\frac{n^{\mathrm{f}}_u}{n^{\mathrm{r}}_u}\right).
\end{equation*}
MissRate counts every IO \textsc{pass} as a miss; censoring instead removes that episode from later risk sets in $\widehat{S}$.
At $H=T_{\max}$ the estimate never exceeds MissRate, because each \textsc{pass} removes an episode from every later risk set and so shrinks every later factor; the two agree only when no \textsc{pass} precedes a later flag.
The detector passes on histories it judges benign, so the censoring is informative, and a common horizon does not remove that dependence. The estimate is not an unbiased evasion probability under continued observation, so we report Survival@$H$ as a descriptive diagnostic next to the terminal error rates.

\section{Implementation Details}
\label{sec:impl_details}
\input{figs/tab_reattack}

\subsection{Agent Architecture}
\label{sec:agent_arch_app}
We parameterize the policy $\pi_g$ as a six-block Mamba-2 sequence model~\citep{dao2024transformers} with width $256$ and state size $64$ over the history $x_{1:t}$.
Each post is represented by one input token, formed by concatenating learned embeddings of its GoEmotions label, posting type, $\log_{10}$ inter-post interval and received engagement counts with a mean-pooled text embedding from the frozen Qwen3.5-4B hybrid encoder\footnote{\url{https://huggingface.co/lylogummy/anima2b-qwen-3.5-4b}} ($2560$-dimensional), and projecting the result to the model width with a trained linear layer.
The backbone maps the token sequence to a summary $h_t$, and three output heads parameterize the factors of Equation~\eqref{eq:autoregressive}.
The emotion head $\pi_e$ is conditioned on $h_t$, the posting-type head $\pi_c$ on $h_t$ and the sampled emotion $e_t$, and the timing head $\pi_\tau$ on $h_t$, $e_t$ and the sampled posting type $c_t$.
During behavioral cloning and preference optimization we train the heads with teacher forcing, conditioning each head on the observed components; during rollouts we sample the three components in sequence and run the backbone in recurrent mode.
A separate critic with the same backbone architecture takes the state $s_t$ as input.
The agent and the detector share an architecture class but no parameters.

\subsection{Learning Rule Details}
\label{sec:ppo_app}
Both players train with the clipped PPO surrogate~\citep{schulman2017proximal}. For the agent, with parameters $\theta_g$, the probability ratio at step $t$ is
\begin{equation*}
w_t(\theta_g)=\frac{\pi_g(a_t\mid s_t;\theta_g)}{\pi_g(a_t\mid s_t;\theta_g^{\text{old}})},
\end{equation*}
and the detector's ratio $w_t(\theta_d)$ is defined the same way with $a^d_t$ and $z_t$. On its turn, player $p\in\{g,d\}$, with policy parameters $\theta_p$ and critic parameters $\psi_p$, maximizes
\begin{equation}
\begin{aligned}
\mathcal{J}_p(\theta_p,\psi_p) ={}& \mathbb{E}_t\Big[\min\big(w_t\hat{A}_t,\;\mathrm{clip}(w_t,1-\epsilon_{\text{clip}},1+\epsilon_{\text{clip}})\,\hat{A}_t\big)\Big] \\
&-c_{\text{val}}\,\mathbb{E}_t\big[(V_{p,t}-V^{\text{targ}}_t)^2\big]
+c_{\text{ent}}\,\mathbb{E}_t\big[\mathcal{H}_{p,t}\big],
\end{aligned}
\label{eq:ppo_app}
\end{equation}
over $n_{\text{ep}}$ epochs of minibatches from that turn's batch. Here $V_{p,t}$ is the player's critic at step $t$, $V^{\text{targ}}_t$ the bootstrapped return target, $\mathcal{H}_{p,t}$ the entropy of its policy at step $t$ and $\hat{A}_t$ the advantage estimate. In the BC-initialized games the agent's objective also subtracts $\lambda_{\text{KL}}\,\mathbb{E}_t[D_{\mathrm{KL}}(\pi_g(\cdot\mid s_t)\,\|\,\pi_g^{\text{BC}}(\cdot\mid s_t))]$; under Equation~\eqref{eq:autoregressive} this KL is the emotion term plus the expected posting-type and timing terms under the adapted policy. The EM and SD games have no such term. Advantages come from generalized advantage estimation~\citep{schulman2016high} with parameter $\lambda_{\text{GAE}}$ and $V_{p,T+1}=0$ at the terminal step, normalized to zero mean and unit variance within each batch. Each player has its own critic; the agent's reads $s_t$ and the detector's reads $z_t$.

\subsection{Replaying Stored Batches}
\label{sec:replay_app}
Let $\mathcal{T}_i$ be the batch of IO trajectories collected at iteration $i$, and $\{\pi_d^{(j)}\}_{j\le i}$ and $\{\mathcal{T}_j\}_{j\le i}$ the histories that \S\ref{sec:training} samples from. Both include the current iteration, and in BC--Real every stored batch is BC rollouts. 
A stored batch was played against an older detector, whose terminal decision fixed the agent's step $T$ and reward. We still render each IO rollout to the full horizon $T_{\max}$, so any later detector can act at every step of the stored history $x_{\le T_{\max}}$.

\subsection{Warm-up, Schedule and Stopping}
\label{sec:schedule_app}
In each game the detector $\pi_d^{(0)}$ starts from random weights and first takes $N_w$ warm-up iterations against the frozen initial agent, so the agent's first updates face a trained opponent. Co-evolution then runs for at most $N_a\ge2W$ iterations. Within a game we drop the game index from snapshots and batches and write $I_k$ for the number of completed co-evolution iterations in $G_k$. Training stops when the window-mean return of each learning player changes by less than $\epsilon_{\text{ret}}$, and the validation MissRate by less than $\epsilon_m$, between two consecutive windows of $W$ iterations (Algorithm~\ref{alg:ioai}, line~\ref{ln:break}). MissRate is measured every iteration on a fixed within-game validation set of held-out IO rollouts and control trajectories from the training pool; no gradient update uses it. The late-window pair $(\bar{\pi}_g^k,\bar{\pi}_d^k)$ of \S\ref{sec:training} is formed whether or not the stopping rule fires before the cap. At rollout time we draw one snapshot per episode and keep it for the whole episode, so a trajectory never switches policy midway.

\subsection{Engagement Score and DPO Objective}
\label{sec:dpo_app}
The engagement score of an action at a state is the DREAMS prediction for the one-step history it produces, averaged over renderer samples,
\begin{equation}
\begin{aligned}
R_{\text{eng}}(s_t,a_t)&=\frac{1}{n_{\text{txt}}}\sum_{j=1}^{n_{\text{txt}}}\mathrm{DREAMS}\big(x_{\le t}\oplus x^{(j)}_{t+1}\big),\\
x^{(j)}_{t+1}&\sim q_{\text{txt}}(\cdot\mid a_t,\kappa),
\end{aligned}
\label{eq:eng_score_app}
\end{equation}
in predicted reactions per post, the sum of predicted comments, shares and emoji reactions. The game reward $\tilde{R}_{\text{eng}}$ standardizes this score by the mean and standard deviation of $R_{\text{eng}}$ over BC rollouts on the training split and clips the result to $[-1,1]$.

EM minimizes the DPO objective~\citep{rafailov2023direct} over the hard-margin preference set $\mathcal{P}_{\text{eng}}$ of \S\ref{sec:agent},
\begin{equation}
\begin{aligned}
\mathcal{L}_{\text{DPO}}
= -\mathbb{E}_{(s,a^+,a^-)\sim\mathcal{P}_{\text{eng}}}\Big[\log \sigma\Big(
&\beta \log \frac{\pi_g(a^+\mid s)}{\pi_g^{\text{BC}}(a^+\mid s)} \\
&- \beta \log \frac{\pi_g(a^-\mid s)}{\pi_g^{\text{BC}}(a^-\mid s)}\Big)\Big],
\end{aligned}
\label{eq:dpo_app}
\end{equation}
where $\sigma$ is the logistic sigmoid and $\beta>0$ scales the log-likelihood ratios. Every pair in $\mathcal{P}_{\text{eng}}$ is labeled with certainty, so the minimizer of Equation~\eqref{eq:dpo_app} pushes the implicit-reward gap without bound~\citep{azar2024general}.

\section{Additional Validation}
\label{sec:additional_validation}

\subsection{Training Dynamics}
\label{sec:training_dynamics_app}
Figure~\ref{fig:training_dynamics} plots policy drift from the BC anchor and predicted engagement over co-evolution for the ten games. \S\ref{sec:results_games} draws on it for the displacement that engagement pretraining gives EM--Evade.

\input{figs/fig_training_dynamics}

\subsection{Re-Attack Under Matched Budget}
\label{sec:reattack_app}
To test resistance to further adaptation, we freeze each of the four detectors in Table~\ref{tab:operating_metrics_main} and train a fresh Evade agent against it from each initialization (BC, EM, SD), with the regularization that initialization uses in \S\ref{sec:games}, for $N_{\text{re}}=100$ PPO iterations on $n=5$ new seeds that neither game solving nor curriculum training used. 
Table~\ref{tab:reattack} reports the original frozen agent's MissRate ($\mathrm{MissRate}_0$), the fresh agent's final MissRate ($\mathrm{MissRate}_{\text{re}}$) on the held-out evaluation set of \S\ref{sec:experiments}, and $i_{0.5}$, the first iteration at which the fresh agent's curriculum-validation MissRate, averaged over a window of $80$, exceeds $0.5$. A dash denotes no crossing within the budget. 
Full IO-AI has the lowest final miss rate for every initialization ($0.18$--$0.23$) and the smallest rise from $\mathrm{MissRate}_0$ ($6$--$8$ pp, against $9$--$15$ pp for the random-order population). The worst case across initializations is $0.23$ for full IO-AI, $0.37$ for the random-order population, $0.64$ for co-evolution against SD agents and $0.81$ for static training, and no fresh agent crosses $0.5$ against full IO-AI. 
EM-initialized attackers have the lowest initial and final miss rates against the static, random-order population and full IO-AI detectors. The detector co-evolved against SD agents stays specialized, with final miss rates of $0.31$ against SD, $0.64$ against BC and $0.41$ against EM. 
Here EM is the engagement-pretrained start; all fresh agents optimize Evade.

\subsection{Ablations}
\label{sec:ablations}
\textbf{Action-space factorization.}
A fully factorized policy, which picks emotion, posting type and timing independently, evades less than the autoregressive one. Late-window evasion falls by $4.3$ pp on average across the three Evade-only games and by $7.1$ pp in EM--Evade, so part of the evasion comes from how the three components are coordinated.

\textbf{BC anchor.}
Without the KL anchor, the BC-initialized agents evade $5$--$8$ pp more, but the cosine similarity between their generated sequences and the original IO sequences falls from $0.71$ to $0.42$. The anchor buys fidelity to the cloned behavior at the price of some evasion.

\textbf{Text features and rendered controls.}
Two checks ask whether the detector keys on traces of the renderer rather than on behavior. Removing its text features changes each main metric by at most $2$ pp, and so does regenerating the control users' text through $q_{\text{txt}}$ so that both classes pass through the renderer.

%% file: figs/tab_archives.tex
\begin{table*}[t]
\centering
\small
\setlength{\tabcolsep}{4pt}
\caption{Complete archive files used for descriptive checks. RT\% is the retweet share; language codes include \texttt{in} (Indonesian) and \texttt{ar} (Arabic).}
\label{tab:archives}
\begin{tabular}{@{}lllrrrll@{}}
\toprule
Archive & Actor & Span & Tweets & Accts. & RT\% & Top languages & Signature client / hashtags \\
\midrule
\texttt{ira\_tweets}    & Russia / IRA & 2009--2018 & 8{,}768{,}633 & 3{,}608 & 38.0 & ru, en            & \texttt{twitterfeed} (1.47M); \#news, \#maga \\
\texttt{ira\_092020}    & Russia / IRA & 2019--2020 & 1{,}368       & 5       & 36.1 & en, \textbf{ar}   & \texttt{PeaceMonitor}; \#blacklivesmatter \\
\texttt{GRU\_202012}    & Russia / GRU & 2013--2020 & 26{,}684      & 51      & 14.0 & en, ru, ar        & \texttt{westvoicesnews}; \#syria, \#idlib \\
\texttt{china\_082019}  & China        & 2008--2019 & 1{,}898{,}108 & 699     & 18.8 & \textbf{in}, en   & \texttt{IFTTT} (824k); \#ptl, \#openfollow \\
\bottomrule
\end{tabular}
\end{table*}

%% file: figs/tab_op_recall.tex
\begin{table}[tp]
\centering
\setlength{\tabcolsep}{4pt}
\caption{Account-level IO recall at detector-specific thresholds calibrated to $10\%$ false-positive rate on held-out benign controls; means over five seeds. GRU is the static detector's weakest archive ($0.42$) and full IO-AI's largest gain ($+0.31$).}
\label{tab:op_recall}
\begin{tabular}{lrrr}
\toprule
Archive  & Static, 12 campaigns & Full IO-AI & Gain \\
\midrule
IRA 2018 
  & {0.90} & {0.94} & {$+$0.04} \\
IRA 2020 
  & {0.76} & {0.88} & {$+$0.12} \\
China 2019 
  & {0.93} & {0.96} & {$+$0.03} \\
GRU 2020 
  & $0.42$ & $0.73$ & $+0.31$ \\
\bottomrule
\end{tabular}
\end{table}

%% file: figs/tab_reattack.tex
\begin{table*}[tp]
\centering
\small
\setlength{\tabcolsep}{4pt}
\caption{Re-attack under matched budget. A fresh Evade agent is trained for $N_{\text{re}}=100$ PPO iterations ($n=5$ seeds) against each frozen detector. A dash denotes no crossing within the budget.}
\label{tab:reattack}
\begin{tabular}{l ccc ccc ccc}
\toprule
& \multicolumn{3}{c}{BC attacker} & \multicolumn{3}{c}{EM attacker} & \multicolumn{3}{c}{SD attacker} \\
\cmidrule(lr){2-4}\cmidrule(lr){5-7}\cmidrule(lr){8-10}
Frozen detector & $\mathrm{MissRate}_0$ & $\mathrm{MissRate}_{\text{re}}$ & $i_{0.5}$ & $\mathrm{MissRate}_0$ & $\mathrm{MissRate}_{\text{re}}$ & $i_{0.5}$ & $\mathrm{MissRate}_0$ & $\mathrm{MissRate}_{\text{re}}$ & $i_{0.5}$ \\
\midrule
Static, 12 campaigns
  & $0.41{\pm}0.07$ & $0.78{\pm}0.05$ & $18{\pm}4$
  & $0.28{\pm}0.06$ & $0.52{\pm}0.06$ & $41{\pm}7$
  & $0.46{\pm}0.08$ & $0.81{\pm}0.04$ & $14{\pm}3$ \\
Co-evolution, SD agents
  & $0.33{\pm}0.05$ & $0.64{\pm}0.06$ & $29{\pm}5$
  & $0.24{\pm}0.04$ & $0.41{\pm}0.05$ & $67{\pm}9$
  & $0.19{\pm}0.03$ & $0.31{\pm}0.04$ & -- \\
Population, random
  & $0.22{\pm}0.04$ & $0.37{\pm}0.05$ & $72{\pm}8$
  & $0.17{\pm}0.03$ & $0.26{\pm}0.04$ & --
  & $0.21{\pm}0.04$ & $0.34{\pm}0.05$ & $81{\pm}10$ \\
Full IO-AI
  & $0.15{\pm}0.03$ & $0.23{\pm}0.04$ & --
  & $0.12{\pm}0.02$ & $0.18{\pm}0.03$ & --
  & $0.16{\pm}0.03$ & $0.22{\pm}0.03$ & -- \\
\bottomrule
\end{tabular}
\end{table*}

%% file: figs/fig_training_dynamics.tex
\begin{figure}[tp]
\centering
\begin{subfigure}[t]{\columnwidth}
\centering
\includegraphics[width=\linewidth]{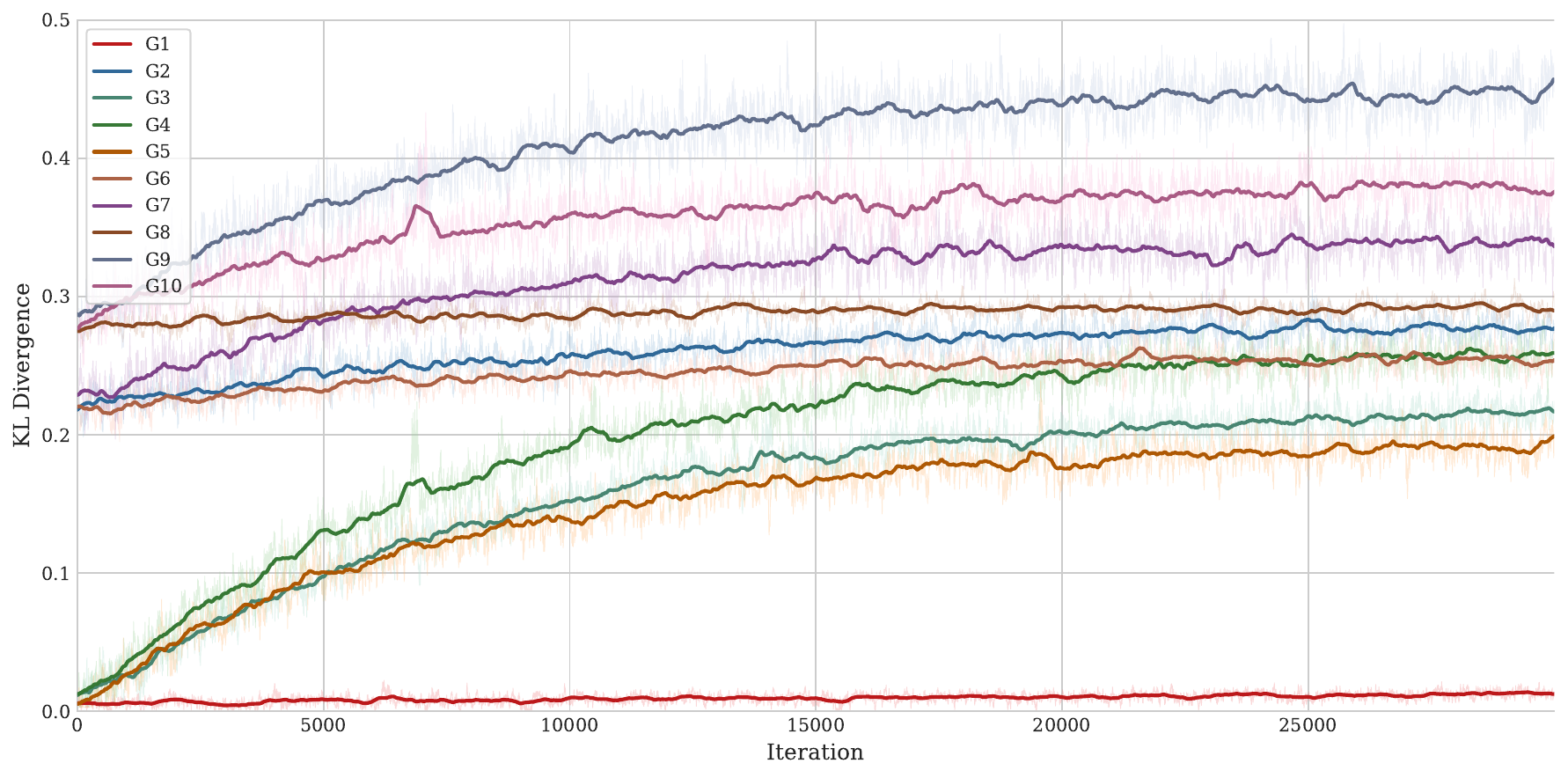}
\caption{Drift}
\label{fig:kl_dynamics}
\end{subfigure}\\[4pt]
\begin{subfigure}[t]{\columnwidth}
\centering
\includegraphics[width=\linewidth]{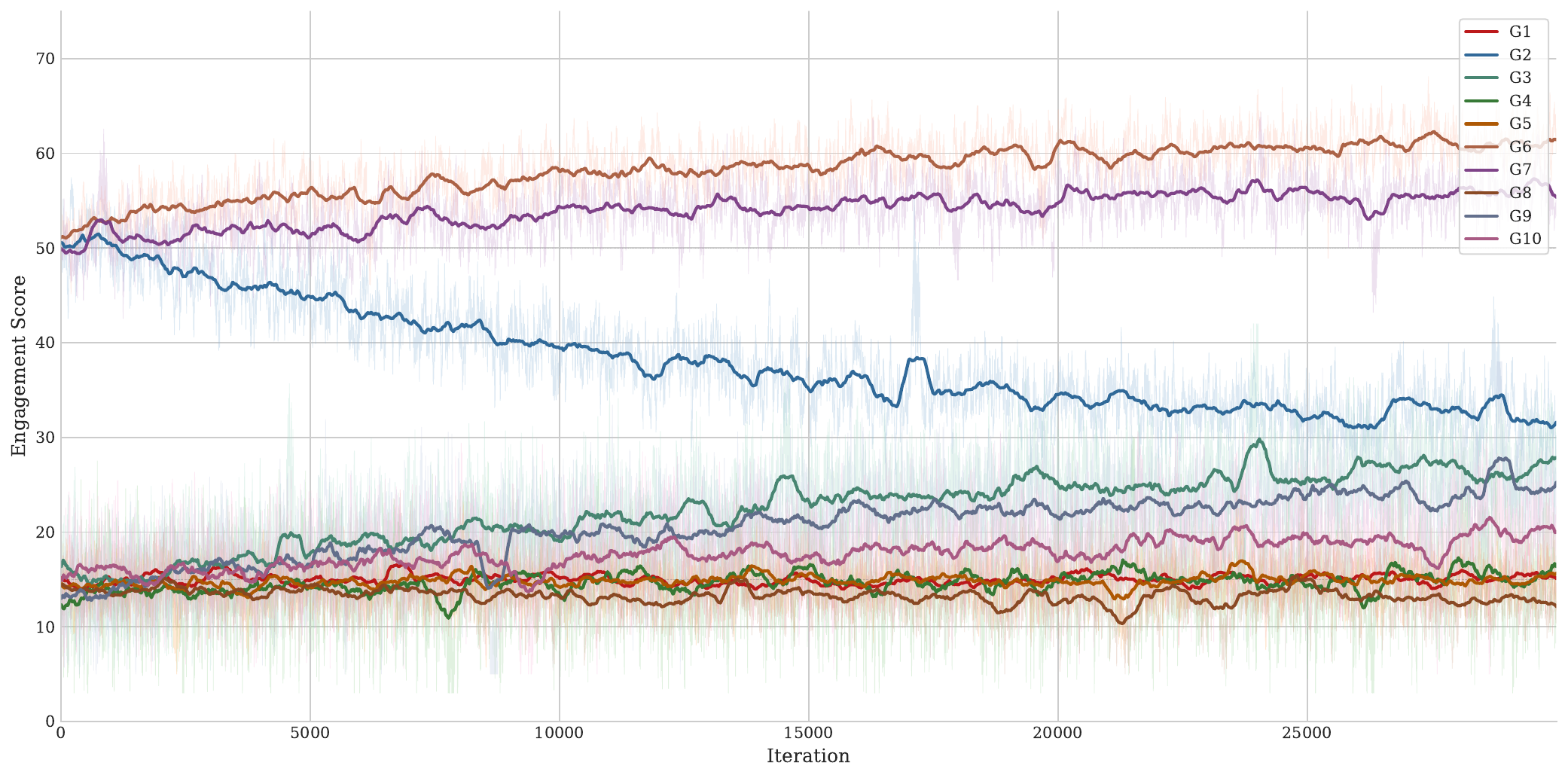}
\caption{Engagement}
\label{fig:engagement_dynamics}
\end{subfigure}
\caption{Training dynamics across the ten games, with (a) policy drift (KL divergence from the BC anchor) and (b) predicted engagement score.}
\label{fig:training_dynamics}
\end{figure}